\documentclass[runningheads]{llncs}

\usepackage[mobile]{eccv} 

\usepackage{eccvabbrv}

\usepackage{amsmath,amsfonts,bm}

\def\figref#1{Fig.~\ref{#1}}

\def\eqref#1{equation~\ref{#1}}

\def\1{\bm{1}}

\def\ve{{\bm{e}}}

\def\vg{{\bm{g}}}

\def\vm{{\bm{m}}}

\def\vp{{\bm{p}}}

\def\vu{{\bm{u}}}
\def\vv{{\bm{v}}}
\def\vw{{\bm{w}}}
\def\vx{{\bm{x}}}
\def\vy{{\bm{y}}}
\def\vz{{\bm{z}}}

\def\mC{{\bm{C}}}

\DeclareMathAlphabet{\mathsfit}{\encodingdefault}{\sfdefault}{m}{sl}
\SetMathAlphabet{\mathsfit}{bold}{\encodingdefault}{\sfdefault}{bx}{n}

\def\gD{{\mathcal{D}}}

\def\gL{{\mathcal{L}}}

\newcommand{\R}{\mathbb{R}}

\usepackage{graphicx}
\usepackage{booktabs}
\usepackage{tcolorbox}
\usepackage{graphbox}
\usepackage{url}
\usepackage{multirow}
\usepackage{booktabs}
\usepackage{colortbl}
\usepackage{tabulary}
\usepackage{algorithmicx}
\usepackage{algpseudocode}
\usepackage{listings}
\usepackage{multicol}
\usepackage{xspace}
\usepackage{wrapfig}
\usepackage{bbm}

\usepackage[accsupp]{axessibility}

\usepackage{kotex}
\usepackage{adjustbox}

\definecolor{Green}{rgb}{0.2, 0.7, 0.1}

\newlength\subsubsecmargin
\newlength\abovetabcapmargin
\newlength\belowtabcapmargin
\newlength\abovefigcapmargin
\newlength\belowfigcapmargin

\usepackage{hyperref}

\usepackage{orcidlink}

\begin{document}

\title{Switch Diffusion Transformer: Synergizing Denoising Tasks with Sparse Mixture-of-Experts} 

\titlerunning{Switch-DiT}

\author{Byeongjun~Park\inst{1}\orcidlink{0000-0002-1930-2266} \and
Hyojun~Go\inst{2}\orcidlink{0000-0002-5470-042X} \and
Jin-Young~Kim\inst{2}\orcidlink{0000-0002-9106-2922} \and
Sangmin~Woo\inst{1}\orcidlink{0000-0003-4451-9675} \and
Seokil~Ham\inst{1}\orcidlink{0000-0003-4400-847X} \and
Changick~Kim\inst{1}\orcidlink{0000-0001-9323-8488}
}

\authorrunning{Park et al.}

\institute{KAIST \and Twelve Labs}

\maketitle

\begin{abstract}

Diffusion models have achieved remarkable success across a range of generative tasks. Recent efforts to enhance diffusion model architectures have reimagined them as a form of multi-task learning, where each task corresponds to a denoising task at a specific noise level. While these efforts have focused on parameter isolation and task routing, they fall short of capturing detailed inter-task relationships and risk losing semantic information, respectively. In response, we introduce Switch Diffusion Transformer (Switch-DiT), which establishes inter-task relationships between conflicting tasks without compromising semantic information. To achieve this, we employ a sparse mixture-of-experts within each transformer block to utilize semantic information and facilitate handling conflicts in tasks through parameter isolation. Also, we propose a diffusion prior loss, encouraging similar tasks to share their denoising paths while isolating conflicting ones. Through these, each transformer block contains a shared expert across all tasks, where the common and task-specific denoising paths enable the diffusion model to construct its beneficial way of synergizing denoising tasks. Extensive experiments validate the effectiveness of our approach in improving both image quality and convergence rate, and further analysis demonstrates that Switch-DiT constructs tailored denoising paths across various generation scenarios. Our project page is available at \href{https://byeongjun-park.github.io/Switch-DiT/}{https://byeongjun-park.github.io/Switch-DiT/}.

\end{abstract}
\section{Introduction}
\label{sec:intro}

Diffusion models have emerged as powerful generative models, showcasing their prowess in various domains, including image~\cite{dhariwal2021diffusion, rombach2022high, peebles2022scalable}, video~\cite{harvey2022flexible} and 3D object~\cite{poole2022dreamfusion, liu2023syncdreamer, woo2023harmonyview}. Specifically, they have made substantial strides across a range of image generation contexts, such as unconditional~\cite{ho2020denoising, song2020denoising}, class-conditional~\cite{dhariwal2021diffusion,nichol2021improved}, and multiple conditions~\cite{brooks2023instructpix2pix}. This progress is attributed to diffusion models learning denoising tasks across various noise distributions, transforming random noise into the desired data distribution through an iterative denoising process.

Given the imperative to learn multiple denoising tasks, recent studies~\cite{Hang_2023_ICCV, go2024addressing} have introduced the concept of multi-task learning (MTL) and revealed that learning denoising tasks leads to the negative transfer between conflicting tasks, resulting in slow convergence of diffusion training. Moreover, they group denoising tasks into three to five clusters based on timestep intervals, demonstrating the effectiveness of training denoising tasks with adjacent timesteps together and separating the learning processes of different denoising task clusters.

These observations align with the improvement shown in architectural design with multiple experts~\cite{lee2023multi, feng2023ernie, balaji2022ediffi, zhang2023improving, xue2024raphael}, wherein denoising tasks are grouped into a small number of clusters, with specialized model parameters assigned to each cluster.
Although their explicit isolation of model parameters according to task clusters has achieved significant performance gain, their manual design of separating conflicted denoising tasks falls short of representing detailed inter-task relationships, and defining which denoising tasks among a thousand possess conflicting optimization directions remains challenging. 

In contrast to isolating model parameters, DTR~\cite{park2023denoising} addresses this issue by constructing distinct data pathways for hundreds of denoising task clusters using predefined task-wise channel masks. This allows diffusion models to develop their own effective way of handling conflicts in denoising tasks.
Additionally, it establishes detailed inter-task relationships, where denoising tasks at adjacent timesteps exhibit a high correlation for smaller timesteps while showing a decreased correlation for larger timesteps. 
However, it's important to note that DTR may lose semantic information due to its channel masking strategy. Consequently, these inherent challenges of previous methods prompt us to inquire:

\begin{tcolorbox}[before skip=0.1cm, after skip=0.1cm, boxsep=0.0cm, middle=0.1cm, top=0.1cm, bottom=0.1cm]
\textit{\textbf{(Q)}} \textit{How to effectively exploit inter-task relationships between conflicted denoising tasks without risking the loss of semantic information?} 
\end{tcolorbox}

In this paper, we delve into the above research question \textit{\textbf{(Q)}} and introduce the Switch Diffusion Transformer \textbf{(Switch-DiT)}, which employs sparse mixture-of-experts (SMoE) layers within each transformer block. In particular, Switch-DiT integrates previous works on MTL-based architectural design. The sparsity in SMoE layers facilitates parameter isolation between conflicted denoising tasks, while inter-task relationships are embodied by its timestep-based gating network. To leverage inter-task relationships, we introduce the diffusion prior loss, which regularizes the output of the gating network. Recognizing that the traditional load-balancing loss~\cite{shazeer2017outrageously} fails to converge the Exponential Moving Average (EMA) model, we instead use the predefined task-wise channel mask in DTR as the strong supervision for the output of the gating network. This design not only exploits inter-task relationships to construct parameter isolation within the diffusion model but also facilitates the rapid convergence of the EMA model.

Furthermore, combining the architectural and loss designs ensures that each transformer block contains at least one shared expert across all denoising tasks, resulting in the construction of common denoising paths and task-specific ones.
This enables our Switch-DiT to forge its own beneficial way of synergizing denoising tasks without compromising semantic information by preserving valuable information within the common denoising path, while task-specific experts capture the remaining information that may contribute to negative transfer.

We conduct experiments on both unconditional~\cite{karras2019style} and class-conditional~\cite{deng2009imagenet} image generation datasets.
Through these experiments, we validate the effectiveness of Switch-DiT in constructing tailored denoising paths for various generation scenarios, thereby improving both image quality and convergence rate.
\section{Related Works}

\subsection{Diffusion model architectures}

Most earlier diffusion models leverage the UNet-based~\cite{ronneberger2015u} and several enhancements have been made within various diffusion model frameworks.
For instance, DDPM~\cite{ho2020denoising} incorporate group normalization and self-attention into the UNet architecture, while IDDPM~\cite{nichol2021improved} advances this design by integrating multi-head self-attention.
ScoreSDE~\cite{song2020denoising} refines the UNet-based architecture by modulating the scale of skip connections, and ADM~\cite{dhariwal2021diffusion} introduces adaptive group normalization to accommodate class-label and timestep embeddings.
More recently, the trend has shifted towards Transformer-based architectures for diffusion models, exemplified by works such as GenViT~\cite{yang2022your}, U-ViT~\cite{bao2023all}, and RIN~\cite{jabri2022scalable}.
Among these, DiT~\cite{peebles2022scalable} stands out by adopting a latent diffusion framework with a Transformer, showcasing notable success.
MDT~\cite{gao2023masked} builds on the DiT framework, further enhancing it with masked latent modeling to better grasp contextual nuances.

While the previously mentioned works primarily leverage a single model to address denoising tasks across various timesteps, a number of studies have investigated the use of multiple expert models, with each specializing in a distinct range of timesteps.
PPAP~\cite{Go_2023_CVPR} implements this by training multiple classifiers on segmented timesteps, each utilized in classifier guidance.
Similarly, e-DiffI~\cite{balaji2022ediffi} and ERNIE-ViLG 2.0~\cite{feng2023ernie} employ a set of denoiser, maintaining consistent architecture across these experts, while MEME~\cite{lee2023multi} presents an argument for the necessity of distinct architecture tailored to each timestep segment. 
Such methodologies enhance generative quality while maintaining comparable inference costs, albeit at the expense of increased memory requirements. 
They rest on the premise that denoising task characteristics vary significantly across timesteps, justifying the deployment of dedicated models.
However, the strict division of parameters among these models could impede beneficial cross-task interactions. 
Our work, therefore, seeks to refine this paradigm, proposing a unified model framework that effectively addresses the spectrum of denoising tasks, fostering and capitalizing on the potential for positive transfer between them.

\subsection{MTL contexts in diffusion model architectures}

Recent studies have reinterpreted diffusion model training through the lens of multi-task learning, where each denoising task at individual timesteps is considered a separate task~\cite{go2024addressing, Hang_2023_ICCV}. 
This perspective has led to the identification of the negative transfer phenomenon, where the conventional multi-task framework can inadvertently compromise denoising performance~\cite{go2024addressing}. 
Differing from this, DTR~\cite{park2023denoising} leverages the inherent multi-task structure of diffusion training to mitigate negative transfer, introducing task-specific information pathways within a unified network architecture. 
Building upon this foundation, our research aims to enhance this architectural strategy, optimizing it to not only minimize negative transfer but also to foster positive interactions among the various denoising tasks, thereby enhancing overall model efficacy.

\subsection{Mixture-of-Experts}

The Mixture-of-Experts (MoE) architecture employs a variety of sub-models and enables conditional computation~\cite{jacobs1991adaptive, shazeer2017outrageously}, proving to be a significant approach in both computational efficiency and model scalability.
In recent developments, MoE methodologies have been instrumental in reducing computational demands during inference, allowing the training of exceptionally large models, such as those with trillions of parameters, particularly within the NLP field~\cite{fedus2022switch, lepikhin2020gshard}.
The approach has also seen successful applications in computer vision, underscoring its versatility across domains~\cite{riquelme2021scaling, wu2022residual}.
A critical aspect of MoE systems is the routing algorithm, which has been the focus of extensive research aimed at enhancement.
This includes a range of routing strategies, such as token-based expert selection~\cite{shazeer2017outrageously,lepikhin2020gshard,fedus2022switch,hazimeh2021dselect}, static routing~\cite{roller2021hash}, and expert-centric token selection~\cite{zhou2022mixture}.
Moreover, several innovative routing algorithms have incorporated auxiliary losses~\cite{shazeer2017outrageously, lewis2021base,clark2022unified} to ensure balanced token distribution across experts, employing sophisticated methods like linear assignment~\cite{lewis2021base} and optimal transport~\cite{clark2022unified} to optimize routing efficiency.

In the diffusion literature, a text-to-image generation method RAPHAEL~\cite{xue2024raphael} employs MoEs for text token and timestep embedding, with its gating networks routing one expert in each MoE layer. This approach fails to leverage inter-task relationships and is incompatible with efficient diffusion learning, as it potentially leads to mode collapse or slowing down convergence due to balancing issues. In contrast, we have devised a sparse MoE-based routing algorithm and incorporated auxiliary loss to model inter-task relationships and serve strong supervision to the gating network, improving both the image quality and convergence speed.
\section{Preliminary}

\subsubsection{Diffusion models.} 

In diffusion models~\cite{dhariwal2021diffusion,song2020denoising}, data is stochastically processed from its initial state ${\vx}_0$ through a sequence of noise addition steps to produce latent representations, which typically follow a Gaussian distribution, in a procedure known as the forward process. Conversely, the reverse process aims to reconstruct the data's original form from its latent state, thereby modeling the data's original distribution $p({\vx}_0)$. The forward process is described as a Markov chain with $T$ steps, during which the data, at each step $t$, transitions according to a Gaussian conditional distribution $q({\vx}_{1:T}|{\vx}_0)$, specifically $q({\vx}_t|{\vx}_0) = \mathcal{N}({\vx}_t; \sqrt{\bar{\alpha}_t}{\vx}_0, (1 - \bar{\alpha}_t)\mathbf{I})$, with $\bar{\alpha}_t$ indicating the noise level. The reverse process entails estimating $p({\vx}_{t-1}|{\vx}_t)$ to approximate $q({\vx}_{t-1}|{\vx}_t)$ and reverse the diffusion to obtain the initial data from its noisy counterpart. A common training approach in diffusion models involves minimizing an objective function as per DDPM~\cite{ho2020denoising} to refine a noise prediction network ${\bm{\epsilon}}_{\bm{\theta}}({\vx}_t, t)$, represented by $\sum_{t=1}^{T} \mathcal{L}_{noise, t}$ where $\mathcal{L}_{noise, t}$ is the expectation of the squared L2 norm difference between the actual and predicted noise, expressed as:
\begin{equation}
\label{eq:simple_loss}
\mathcal{L}_{noise, t}:=\mathbb{E}_{{\vx}_0, {\bm{\epsilon} \sim \mathcal{N}(0, 1)}} \| {\bm{\epsilon}} - {\bm{\epsilon}}_{\bm{\theta}}({\vx}_t, t) \|_2^2.
\end{equation}

\subsubsection{Sparse Mixture-of-Experts.} In general, MoE layers~\cite{shazeer2017outrageously, zoph2022st} contain $M$ expert networks $E_1, E_2, \dots, E_M$ and a gating network $G$. Both expert networks and the gating network often consist of a single MLP layer and take the same input $x$. The output of each MoE layer is the weighted sum of the expert network outputs and gating network outputs, which is formulated as: 
\begin{equation}
y =\sum_{i=1}^{M} G^i(x)  E_{i}({x}),
\end{equation}
where $G^i(x) \in \R$ presents the weight for $i$-th expert network. Sparse MoE layers avoid computing on all expert networks by selecting the largest $k$ elements of the gating network outputs. To this end, the gating network is re-defined as:
\begin{equation}
    G(x) = \operatorname{TopK}(\operatorname{Softmax} (h(x) + \epsilon), k),
\end{equation}
where an MLP layer $h$ and trainable noise $\epsilon$ are incorporated. For sparsity, $\operatorname{TopK}(\cdot,k)$ sets all elements into zero except the largest $k$ elements.
\section{Methods}

In this section, we present a Switch Diffusion Transformer (Switch-DiT), a novel diffusion model architecture embodying a sparse mixture-of-experts (SMoE) in each transformer block.
We aim to synergize multiple denoising tasks during diffusion training by preserving valuable information within a shared denoising path while isolating model parameters to handle task-specific information that may lead to negative transfer.
We begin by elucidating the design space of Switch-DiT in Sec.~\ref{subsec:design_space}, which encompasses timestep-based gating networks, integration of SMoE layers with transformer blocks, and configuration for setting the number of experts and TopK values. Additionally, we discuss the diffusion prior loss in Sec.~\ref{subsec:aux_loss}, which stabilizes the convergence of the EMA model for gating networks and encourages similar tasks to share denoising paths while regularizing conflicting tasks to take distinct denoising paths.

\begin{figure*}[t]
    \centering
    \includegraphics[width=\linewidth]{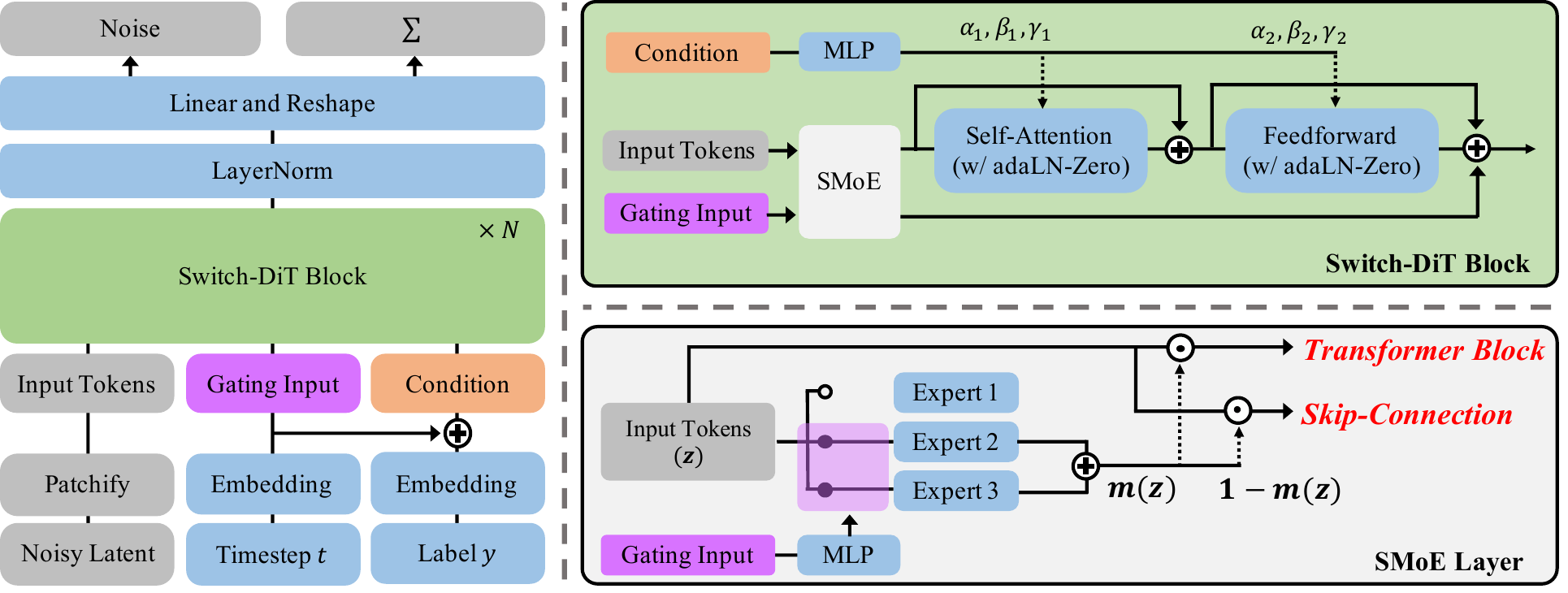}

    \caption{\textbf{Switch Diffusion Transformer.} $\odot$ represents an element-wise multiplication. Switch-DiT is built upon the DiT~\cite{peebles2022scalable} architecture, which consists of the self-attention and the feedforward, both conditioned on timestep embeddings and label embeddings via the adaLN-Zero layer. In the SMoE layer, the gating network takes the timestep embeddings and selects two out of three experts to output $\vm(\vz)$. Then, $\vz \cdot \vm(\vz)$ is used as input to the transformer block, and $\vz \cdot (1 - \vm(\vz))$ is skip-connected to the end.}

    \label{fig:switch-dit}
\end{figure*}
\subsection{Switch-DiT Design Space}
\label{subsec:design_space}

Since our Switch-DiT is built upon the DiT~\cite{peebles2022scalable} architecture as shown in~\figref{fig:switch-dit}, we start by delineating its components. DiT comprises $N$ transformer blocks, where the $i$-th transformer block takes as inputs the timestep embedding $\ve_t$ for discrete timestep $t \in \{1, \dots, T\}$, label embedding $\vy$ and input token $\vz_{i}$, and produces the output $\vz_{i+1}$. Subsequently, the final output of the transformer blocks $\vz_{N}$ is processed to estimate the noise added to the latent representation.

\subsubsection{Timestep-based gating network.} In contrast to traditional gating networks that use the same inputs as experts, Switch-DiT employs timestep embeddings $\ve_t$ as the input of the gating network, focusing on parameter isolation based on denoising tasks. Therefore, our gating outputs in $i$-th block $\vg_{i}$ is formulated as:
\begin{equation}
\vg_{i}(\ve_t)=\operatorname{TopK}(\vp_i(\ve_t), k),
\end{equation}
where $\vp_i(\ve_t)$ is the probability of each expert being selected for timestep $t$ as:
\begin{equation}
\vp_i(\ve_t)=\operatorname{Softmax} (h_{i}(\ve_t)).
\end{equation}

Here, $h_{i}$ is an MLP layer. To enforce sparsity, we also use $\operatorname{TopK}(\cdot, k)$ to set all elements into zero except the largest $k$ elements. Note that we utilize a straightforward gating function and refrain from employing Noisy TopK Gating~\cite{shazeer2017outrageously}, which is often coupled with load-balancing loss or z-loss~\cite{zoph2022st} to ensure balanced token distribution among experts. This design choice arises from its unsuitability for diffusion training, notably the failure of the Exponential Moving Average (EMA) model of the gating network to converge. This causes the TopK selection process to be inconsistent with the intended behavior during diffusion training.

\subsubsection{SMoE layer design.} Following the typical mixture-of-experts methods~\cite{shazeer2017outrageously, zoph2022st, fedus2022switch}, we use an MLP layer for each expert, where $M$ experts in $i$-th block $E^1_i, E^2_i, \dots, E^M_i$ and the gating outputs constitute the SMoE layer as follows:
\begin{equation}
\vm(\vz_{i})=\sum_{j=1}^{M} \vg_{i, j}(\ve_t) E_{i}^{j}({z_{i}}),
\end{equation}
where we do not activate experts whose gating output has zero probability.

For the output of the SMoE layer $\vm(\vz_{i})$, we explore three variants of the integration with a transformer block. First, we simply use $\vm(\vz_{i})$ as the input of the remaining transformer block (\ie, original DiT block). Second, we utilize the same as the task-wise channel mask in DTR~\cite{park2023denoising}, where input tokens are element-wise multiplied to be used as input to the transformer block, and the residual term $\vz_{i} \cdot (1-\vm(\vz_{i}))$ is skip-connected at the end. Finally, we adhere to prior practices which initialize any residual blocks as the identity  function~\cite{goyal2017accurate, dhariwal2021diffusion, peebles2022scalable}, thus each SMoE output is initialized as a one-vector in addition to the second design. We have empirically found that the last design performs best, which allows the SMoE layer at each transformer block to have minimal impact at the beginning of the training, while the diffusion model can learn how to effectively synergize denoising tasks on its own during the training procedure.

\subsubsection{Number of experts and TopK values.} For each transformer block, our SMoE layer aims to establish a shared denoising path across all denoising tasks while effectively isolating model parameters between conflicting tasks. Specifically, to facilitate the construction of both a shared denoising path and task-specific paths, the TopK value ($k$) must be at least two and the number of experts ($M$) has to be greater than $k$ for the sparsity. Therefore, we choose the most efficient configuration with $M=3$ and $k=2$ to meet these requirements.

\subsection{Auxiliary Loss Design}
\label{subsec:aux_loss}

\begin{figure*}[t]
    \centering
    \includegraphics[width=\linewidth]{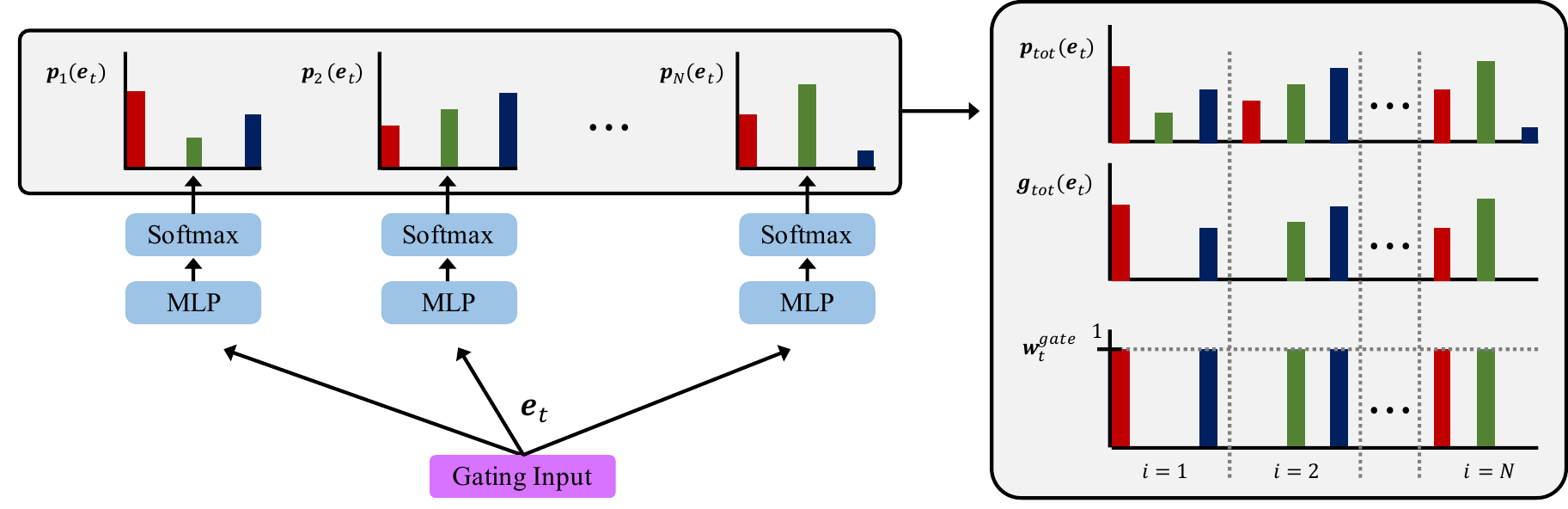}
    
    \caption{\textbf{Gating Outputs Integration.} For simplicity, we visualize the gating outputs for three experts and select the largest two elements within each transformer block. As a result, $\vp_{tot}(\ve_{t})$ is a concatenated probability of each $\vp_{i}(\ve_{t})$ for $i$-th block, which is then used for the diffusion prior loss. Also, $w_{t}^{gate}$ is used for a cost function of the bipartite matching with that similarly derived from the DTR~\cite{park2023denoising}.}

    \label{fig:integration_gating}
\end{figure*}

Instead of relying on previous load-balancing techniques, which are unsuitable for diffusion training, we introduce a novel auxiliary loss called the diffusion prior loss.  This loss function serves to stabilize the convergence of the EMA model for the gating network and also regularizes the gating outputs to reflect detailed inter-task relationships between denoising tasks. By learning these inter-task relationships, similar denoising tasks can share their denoising paths, facilitating the construction of a shared denoising path across all denoising tasks.

\subsubsection{Gating outputs integration.} We first aggregate $\vp_{i}(\ve_t) \in \R^{M}$ across $N$ transformer blocks, wherein the integrated probability $\vp_{tot}(\ve_t) \in \R^{NM}$ is then utilized to regress the inter-task relationship within the diffusion model. We also aggregate each gating output $\vg_{i}(\ve_t)$ to obtain $\vg_{tot}(\ve_t)$, which is then processed to a task-expert activation map $\vw^{gate}_{t}=\mathbbm{1}[\vg_{tot}(\ve_t) > 0]$. This process sequence is illustrated in \figref{fig:integration_gating}. We utilize the task-expert activation map for the bipartite matching, while the integrated probability is employed for the diffusion prior loss, which is combined with the bipartite matching.
 
\subsubsection{Bipartite matching.} To fully leverage the inter-task relationships described in DTR~\cite{park2023denoising}, we define another task-expert activation map ${\vw}_{t}^{prior} \in \R^{NM}$ derived from its task-wise channel mask. Here, we set $\alpha$ to four and the channel dimension to $NM$ with a sharing ratio of $k/M$. This configuration defines ${\vw}_{t}^{prior}$ as:
\begin{equation}
\label{eq:w_prior}
{\vw}_{t, c}^{prior} = \begin{cases}
  1,                    & \mbox{if $\lfloor N(M - k) \cdot \left(\frac{t-1}{T}\right)^{\alpha}\rceil < c \leq \lfloor N(M - k) \cdot \left(\frac{t}{T}\right)^{\alpha}\rceil + kN$}, \\
  0,                    & \mbox{otherwise}.
\end{cases}
\end{equation}

\begin{wrapfigure}[11]{r}{0.42\textwidth}
    \centering
    \vspace{-8mm}
    \includegraphics[width=\linewidth]{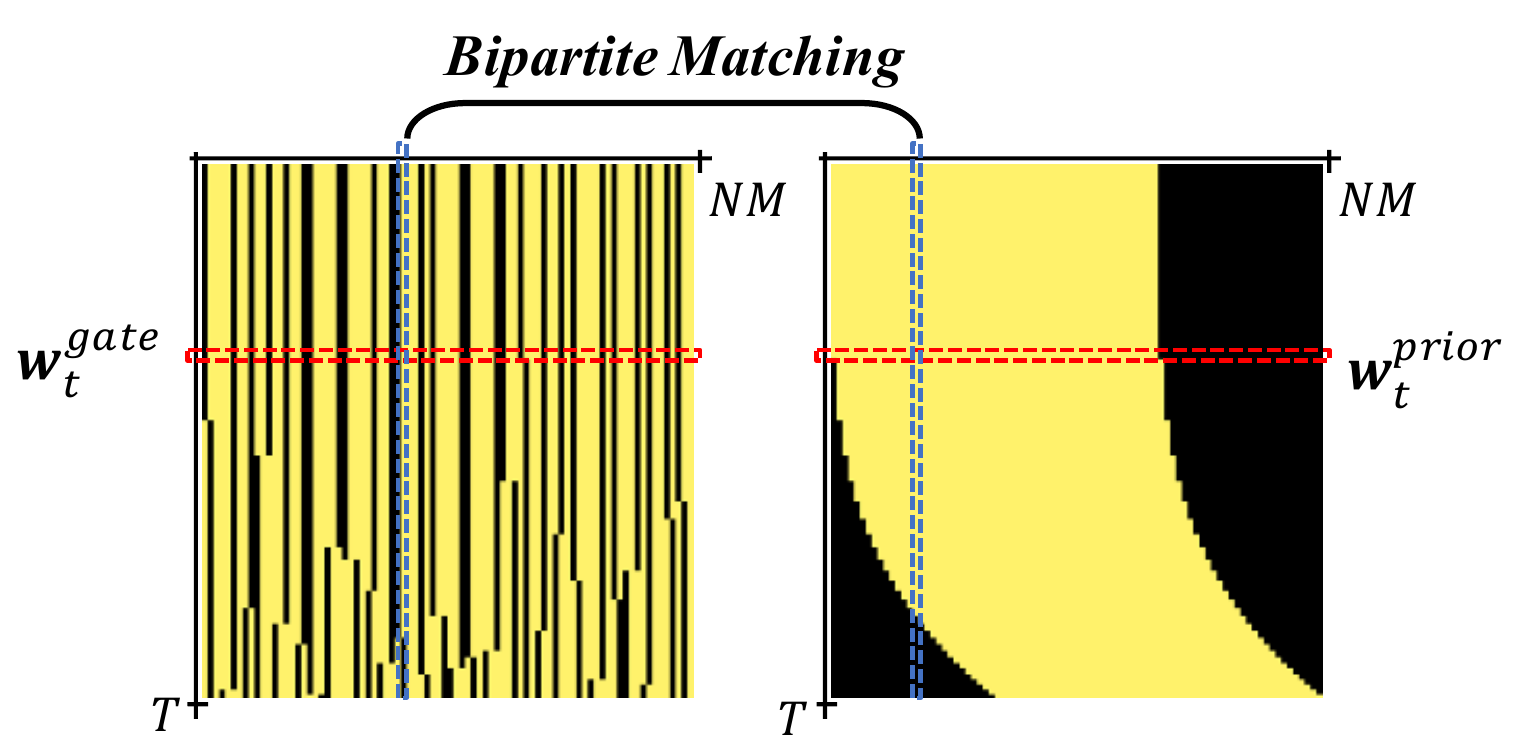}
    \vspace{-7.5mm}
    \caption{\textbf{Bipartite Matching.} We show the stacked $\vw^{gate}_{t}$ and $\vw^{prior}_{t}$ for $N=24$, $M=3$ and $k=2$. Each row represents a concatenated activation map as shown in Fig.~\ref{fig:integration_gating}.
    }
    \label{fig:bipartite_matching}
\end{wrapfigure}


We recap that $\vw^{gate}_{t}$ is a block-wise concatenated binary map, whereas $\vw^{prior}_{t}$ is constructed in a channel shift fashion.
Recognizing the inconsistency in channel permutations of two types of task-expert activation maps, we align them to minimize the cost function using bipartite matching.
Specifically, we stack each activation map $\vw^{gate}_{t}$ and $\vw^{prior}_{t}$ for $T$ timesteps to ensure stable matching, as illustrated in \figref{fig:bipartite_matching}. Then, we define a cost function $\mC \in \R^{NM \times NM}$ as the sum of the pair-wise distance for two activation maps across all timesteps, and $\mC$ is formulated as:
\begin{equation}
    \mC = \sum_{t=1}^{T} \operatorname{cdist}(\vw^{gate}_{t}, \vw^{prior}_{t}),
\end{equation}
and the element at position $(i, j)$ of $\operatorname{cdist}(\vu, \vv)$ is defined as:
\begin{equation}
    \operatorname{cdist}(\vu, \vv)_{ij} = \lVert \vu_i - \vv_j \rVert_{1}.
\end{equation}

Through this cost function and bipartite matching, we can align the channel permutations of $\vw^{gate}_{t}$ to represent best the inter-task relationships derived from the DTR across all timesteps.

\subsubsection{Diffusion prior loss.} After we find the best channel permutation for the gating outputs, we apply this on $\vp_{tot}(\ve_{t})$ to output $\tilde{\vp}_{tot}(\ve_{t})$. We then compute a diffusion prior loss $L_{dp, t}$ for timestep $t$ as a Jensen-Shannon Divergence (JSD):
\begin{equation}
    \gL_{dp, t} = \gD_{JS} \Big( \frac{\tilde{\vp}_{tot}(\ve_{t})}{N} \big\| \frac{\vw^{prior}_{t}}{kN} \Big),
\end{equation}
where the denominators serve as scale factors to ensure that each probability vector sums to one. By minimizing this loss, we ensure that the largest $k$ elements of the gating output are assigned a value of $1/k$, while the unselected elements are guided to output zero probability. This strict regularization facilitates rapid convergence of the EMA model of the gating network and reflects inter-task relationships, making similar tasks share a denoising path. Combined with the noise prediction loss in Eq.~\ref{eq:simple_loss}, each denoising task of the Switch-DiT at timestep $t$ is trained with the weighted sum of $\gL_{noise, t}$ and $\gL_{dp, t}$ as:
\begin{equation}
    \gL_{t} = \gL_{noise, t} + \lambda_{dp} \gL_{dp, t}
\end{equation}

\section{Experiments}

In this section. we begin by outlining our experimental setups in Sec.~\ref{subsec:setup}. Next, we present the experimental results of the Switch-DiT design space in Sec.~\ref{subsec:result_design_space}, validating the effectiveness of Switch-DiT in isolating model parameters without losing semantic information while providing common denoising paths across all tasks. Then, we provide comparative results with MTL-based diffusion models in Sec.~\ref{subsec:comparable_results}. Finally, we conduct a thorough analysis of Switch-DiT in Sec.~\ref{subsec:analysis}.

\subsection{Experimental Setup}
\label{subsec:setup}

\subsubsection{Datasets.}
We conducted experiments on two image-generation tasks. Firstly, we explored the Switch-DiT design space in the unconditional generation task using the FFHQ dataset~\cite{karras2019style}, which consists of 70K human face images.
Second, we verified the effectiveness of the Switch-DiT in the class-conditional generation task using ImageNet~\cite{deng2009imagenet}, which consists of 1,281,167 images for 1K classes. In this experiment, we utilized images with a fixed resolution of $256\times256$.

\subsubsection{Implementation details.}

We utilized a VAE encoder/decoder sourced from Stable Diffusion\footnote{\href{https://huggingface.co/stabilityai/sd-vae-ft-ema-original}{https://huggingface.co/stabilityai/sd-vae-ft-ema-original}}. These are used to extract the latent features from input images and decode the denoised latent features during the sampling process, respectively.
Specifically, the latent representation used by our Switch-DiT has dimensions of $32\times32\times4$, which is derived from $256\times256\times3$ images.
We set the patch size to two to patchify the latent representation.
Following previous methods~\cite{peebles2022scalable, park2023denoising}, we employed an exponential moving average (EMA) on the model parameters during training, utilizing a decay rate of 0.9999 to enhance stability. Then, we used this EMA model during the sampling process.
We set the diffusion timestep $T$ to 1000 and DDPM~\cite{ho2020denoising} steps to 250 for the sampling.
We used a cosine scheduling strategy~\cite{nichol2021improved} and utilized classifier-free guidance~\cite{ho2021classifierfree} with a guidance scale of 1.5 in the class-conditional image generation.

\subsubsection{Training.}
We employed the AdamW optimizer~\cite{loshchilov2017decoupled} with a learning rate of 1e-4 and no weight decay. Also, we applied random horizontal flips and used $\lambda_{dp} = 1$. We trained for 100k and 400k iterations for the FFHQ~\cite{karras2019style} and ImageNet~\cite{deng2009imagenet} datasets, respectively. All models were trained with a batch size of 256 on 8 NVIDIA A100 GPUs.

\subsubsection{Evaluation.}
We report the performance of diffusion models using FID~\cite{heusel2017gans}, IS~\cite{salimans2016improved}, and Precision/Recall~\cite{kynkaanniemi2019improved}.
We followed the ADM's evaluation protocol from its codebase\footnote{\href{https://github.com/openai/guided-diffusion/tree/main/evaluations}{https://github.com/openai/guided-diffusion/tree/main/evaluations}} and reported these metrics from 50K generated samples for all experiments unless otherwise noted as FID-10K evaluated from 10K images.

\begin{table*}[t]
    \begin{center}
    \setlength\tabcolsep{3pt}
    \begin{small}
    \scalebox{0.68}{
    \begin{tabular}{lcccc}
    \toprule
    \multicolumn{5}{l}{\textbf{Timestep-based Gating Network}} \\
    \toprule
    \multirow{2}{*}{Model} & \multicolumn{3}{c}{Design Space} & \multirow{2}{*}{FID$\downarrow$} \\
    \arrayrulecolor{gray}\cmidrule(lr){2-4}
    & Noisy & $L_{load}$ & $L_{dp}$  &  \\
    \arrayrulecolor{black}\midrule 
    DiT-B~\cite{peebles2022scalable}     & & &         &  10.99 \\
    \arrayrulecolor{gray}\cmidrule(lr){1-5}
    \multirow{5}{*}{Switch-DiT-B} & &  &  &  9.68 \\
     & \checkmark &  &  &  8.26 \\
     & \checkmark & \checkmark &  &  16.78 \\
     & \checkmark &  & \checkmark &  7.59 \\
     & \cellcolor{gray!25} & \cellcolor{gray!25} & \cellcolor{gray!25}\checkmark & \cellcolor{gray!25}\textbf{7.12} \\
    \arrayrulecolor{black}\bottomrule
    \end{tabular}
    }
    \scalebox{0.68}{
    \begin{tabular}{lcccc}
    \toprule
    \multicolumn{5}{l}{\textbf{SMoE Layer Integration}}  \\
    \toprule
    \multirow{2}{*}{Model} & \multicolumn{3}{c}{Design Space} & \multirow{2}{*}{FID$\downarrow$} \\
    \arrayrulecolor{gray}\cmidrule(lr){2-4}
    & SMoE & Skip & Init &  \\
    \arrayrulecolor{black}\midrule 
    DiT-B~\cite{peebles2022scalable}    & &   &      &  10.99 \\
    \arrayrulecolor{gray}\cmidrule(lr){1-5}
    \multirow{5}{*}{Switch-DiT-B} & Direct & & & 8.15 \\
     & Mask & &  & 7.98 \\
     & Mask & \checkmark &  &  7.77 \\
     & Mask &  & \checkmark &  7.53 \\
     & \cellcolor{gray!25}Mask & \cellcolor{gray!25}\checkmark & \cellcolor{gray!25}\checkmark & \cellcolor{gray!25}\textbf{7.12} \\
    \arrayrulecolor{black}\bottomrule
    \end{tabular}
    }
    \scalebox{0.68}{
    \begin{tabular}{lcccc}
    \toprule
    \multicolumn{4}{l}{\textbf{Hyper-parameters}}  \\
    \toprule
    \multirow{2}{*}{Model} & \multicolumn{3}{c}{Design Space} & \multirow{2}{*}{FID$\downarrow$} \\
    \arrayrulecolor{gray}\cmidrule(lr){2-4}
    & $M$ & $k$ &  Params &  \\
    \arrayrulecolor{black}\midrule 
    DiT-B~\cite{peebles2022scalable}     & & &  131M       &  10.99 \\
    \arrayrulecolor{gray}\cmidrule(lr){1-5}
    \multirow{5}{*}{Switch-DiT-B} & 3 & 1 & 137M & 8.20 \\
     & 4 & 2 & 144M & 7.83 \\
     & 2 & 2 & 144M & 9.24 \\
     & 4 & 3 & 151M & \textbf{7.05} \\
     & \cellcolor{gray!25}3 & \cellcolor{gray!25}2 & \cellcolor{gray!25}144M &  \cellcolor{gray!25}7.12 \\
    \arrayrulecolor{black}\bottomrule
    \end{tabular}
    }
    \end{small}
    \end{center}

    \caption{\textbf{Switch-DiT Design on the FFHQ dataset.} We provide design spaces for a timestep-baed gating network, an SMoE layer integration, and hyper-parameters.}

\label{tab:design_space}
\end{table*}

\subsection{Switch-DiT Design}
\label{subsec:result_design_space}

We explore the Switch-DiT design space using the FFHQ dataset. The corresponding results are presented in Table~\ref{tab:design_space}, where we validate the effectiveness of our design choices based on these experimental results.

\subsubsection{Timestep-based gating network.} To validate the simplicity and effectiveness of our timestep-based gating network, we investigate the use of the Noisy TopK Gating and load-balancing loss, which are commonly employed in previous MoE methods. Training Switch-DiT without additional components suffers from mode collapse problem -- \textit{all denoising tasks use the same experts}. This prevents Switch-DiT from modeling parameter isolation and inter-task relationships, and any further improvements over DiT can be attributed to the additional model parameters. We also confirm that applying the Noisy TopK Gating improves performance while applying load-balancing loss significantly degrades performance, even worse than DiT. The reason behind this degradation is the failure of the EMA model of the gating network to converge. Thus, the gating behavior deviates from the trained logic, leading to the expert not being utilized as intended.

In contrast, our diffusion prior loss $L_{dp}$ makes the gating network well establish parameter isolation and inter-task relationships across denoising tasks, leading to significant performance improvements. Furthermore, our observations reveal that the additional application of the Noisy TopK Gating tends to decrease performance. Consequently, we opt for a simple yet efficient gating network design, complemented by the application of the diffusion prior loss.

\begin{table*}[t!]
\begin{minipage}[t!]{0.54\textwidth}
    \setlength\tabcolsep{1.5pt}
    \begin{center}
    \begin{small}
    \scalebox{0.72}{
    \begin{tabular}{lcccccc}
    \toprule
    Model & Params & GFLOPs & FID$\downarrow$  & IS$\uparrow$  & Prec$\uparrow$ & Rec$\uparrow$ \\
    \toprule
    DiT-S~\cite{peebles2022scalable}  & 33M & 6.06 & 44.28 & 32.31 & 0.41 & 0.53 \\
    DTR-S~\cite{park2023denoising}  & 33M & 6.06 & 37.43 & 38.97 & 0.47 & \textbf{0.54} \\
    \rowcolor{gray!25}\textbf{Switch-DiT-S} & 36M & 6.74 & \textbf{33.99} & \textbf{42.99} & \textbf{0.52} & \textbf{0.54} \\
    \arrayrulecolor{gray}\cmidrule(lr){1-7}
    DiT-B~\cite{peebles2022scalable} & 131M & 23.01 & 27.96 & 64.72 & 0.57 & 0.52 \\
    DTR-B~\cite{park2023denoising} & 131M & 23.02 & 16.58 & 87.94 & 0.66 & \textbf{0.53} \\
    \rowcolor{gray!25}\textbf{Switch-DiT-B} & 144M & 26.55 & \textbf{16.21} & \textbf{88.14} & \textbf{0.68} & \textbf{0.53} \\
    \arrayrulecolor{gray}\cmidrule(lr){1-7}
    DiT-XL~\cite{peebles2022scalable} & 675M & 118.64 & 9.40 & 166.83 & 0.77 & \textbf{0.48} \\
    DTR-XL~\cite{park2023denoising} & 675M & 118.69 & 10.85 & 158.45 & 0.73 & 0.47 \\
    \rowcolor{gray!25}\textbf{Switch-DiT-XL} & 749M & 132.72 & \textbf{8.76} & \textbf{169.17} & \textbf{0.79} & \textbf{0.48} \\
    \arrayrulecolor{black}\bottomrule
    \end{tabular}}
    \captionof{table}{\textbf{Comparison across model sizes on ImageNet.} Our Switch-DiT achieves consistent performance improvements.}
    \label{tab:scalability_abl}
    \end{small}
    \end{center}
\end{minipage}
\begin{minipage}[t!]{0.42\textwidth}
    \centering
    \includegraphics[width=0.9\textwidth]{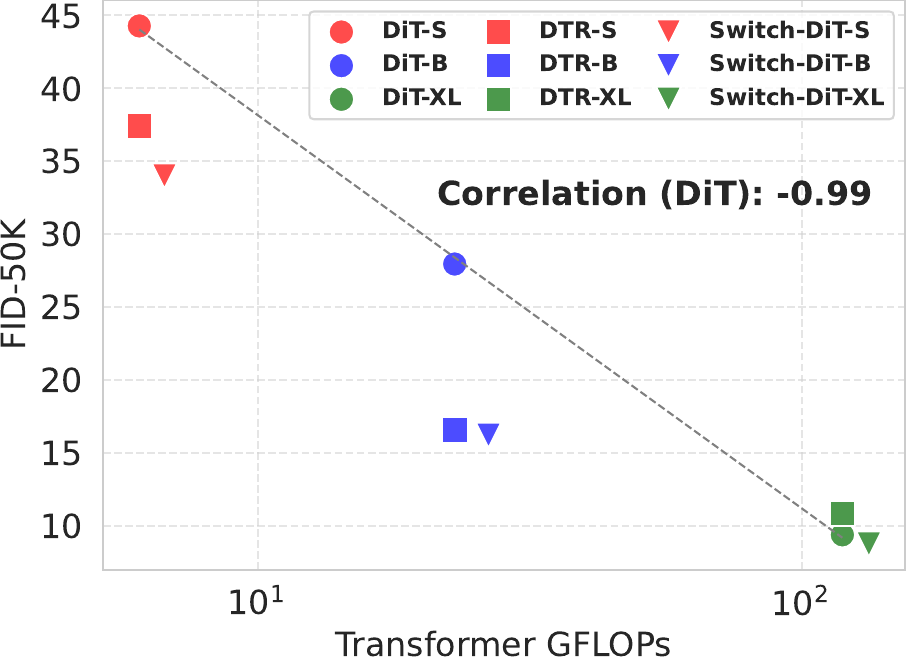}
    \captionof{figure}{\textbf{Correlation of GFLOPs and FID on ImageNet.} Switch-DiT transcends the tradeoff of DiT.}
    \label{fig:correlation}
\end{minipage}
\end{table*}

\subsubsection{SMoE layer integration.} We verify the effectiveness of our design choice in the SMoE layer integration. We confirm that it is more advantageous to utilize $\vz_i \cdot \vm(\vz_i)$ in the subsequent $i$-th transformer block, rather than using $\vm(\vz_i)$ in its original form. We further validate that incorporating the skip-connection of $\vz_i \cdot (1-\vm(\vz_i))$ to the end of the transformer block, as proposed in the DTR, constitutes an additional enhancement. This integration ensures that the residual information from the transformed input tokens through the SMoE layer is effectively utilized throughout the transformer block. Finally, we observe that initializing $\vm(\vz_i)$ as a one-vector leads to better performance, enabling the diffusion model to initiate learning from the identity function. This minimizes the influence of the SMoE layer during the early stages of learning, enabling Switch-DiT to effectively synergize denoising tasks as training progresses.

\subsubsection{Impacts on $M$ and $k$.} We thoroughly delve into the impacts on the number of experts ($M$) and the TopK values ($k$). In the case of $k=1$, there is no common denoising path across all timesteps; rather, exclusive denoising paths are constructed across small clusters of denoising tasks. We confirm the effectiveness of creating both a shared path and a task-specific path by setting $k=2$ 
compared to constructing an exclusive denoising path (FID-10K: 8.20 vs. 7.12). 

We also present results with $M=k$, illustrating that SMoE-based gating logic is more effective while maintaining the same computational complexity during the sampling process (FID-10K: 9.24 vs. 7.12). Interestingly, we observe that increasing $M$ does not always result in performance improvements. This is related to our diffusion prior loss, where the number of shared experts across all timesteps is given by $\max(N \cdot (2k -M), 0)$ in DTR. Therefore, as $M$ increases, the number of shared experts decreases, resulting in no such shared experts in the case of $M=4$ and $k=2$. This highlights the importance of having a sufficient number of common denoising paths. We can verify this with an experimental result for $M=4$ and $k=3$, having more shared denoising paths leads to additional performance improvements. Obviously, there is room for the scalability of our Switch-DiT architecture, but we focus on the synergy of denoising tasks in this paper, thereby our experiments are centered around the most parameter-efficient settings of $M=3$ and $k=2$.

\subsubsection{Relation to time-MoE~\cite{xue2024raphael}.} The time-MoE proposed in RAPHAEL~\cite{xue2024raphael} represents one of our designs, incorporating the Noisy TopK Gating, the direct use of the MoE output, and a configuration set of $M=4$ and $k=1$. Through our design space exploration, we can explain the superiority of Swith-DiT compared to the time-MoE as follows: (1) Establishing inter-task relationships between denoising tasks, (2) More efficient integration with transformer blocks, (3) Synergizing denoising tasks by providing shared denoising paths across all timesteps.

\subsection{Comparative Evaluation}
\label{subsec:comparable_results}

\subsubsection{Quantitative results.}

As shown in Table~\ref{tab:scalability_abl}, we validate the effectiveness of Switch-DiT across different model sizes on the ImageNet dataset. We confirm that Switch-DiT consistently demonstrates better performance improvement compared to DTR. In addition, we observe that DTR-XL shows inferior performance compared to DiT-XL. This underperformance can be attributed to its channel masking, where masking 20\% of the channels to zero leads to a substantial loss of semantic information. This loss may outweigh the benefits of efficient diffusion training in large-scale diffusion models, such as DTR-XL. In contrast, Switch-DiT-XL achieves superior performance compared to both DiT-XL and DTR-XL by effectively handling conflicting tasks through parameter isolation, despite using the same routing policy as DTR.

We acknowledge the potential interpretation of the performance improvement in Switch-DiT as being attributed to additional parameters and the resulting increase in GFLOPs. 
However, given the strong correlation between GFLOPs and FID scores observed in DiT, as well as the lack of correlation between model parameters and FID scores, our Switch-DiT surpasses its tradeoff as shown in \figref{fig:correlation}, demonstrating that the performance enhancement is not solely due to additional parameters and GFLOPs. In particular, according to the correlation, the FID-50K score of DiT-based architecture corresponding to Switch-DiT-S's GLFOPs of 6.74 should be 42.79, whereas the actual FID-50K score of Switch-DiT-S is 33.99, indicating a significant improvement over the expected score.

\begin{wraptable}[11]{r}{0.33\textwidth}
    \setlength\tabcolsep{4pt}
    \centering
    \resizebox{0.33\textwidth}{!}{%
    \begin{tabular}{lc}
    \toprule
    Method & FID-10K$\downarrow$ \\
    \midrule
    DiT-B~\cite{peebles2022scalable} & 12.93 \\
    \arrayrulecolor{gray}\cmidrule(lr){1-2}
    \multicolumn{2}{l}{\textit{Loss Design}} \\ 
    \quad $+$ P2~\cite{choi2022perception} & 10.08 \\
    \quad $+$ Min-SNR~\cite{Hang_2023_ICCV} & 9.73 \\
    \quad $+$ ANT-UW~\cite{go2024addressing} & 9.30 \\
    \arrayrulecolor{gray}\cmidrule(lr){1-2}
    \multicolumn{2}{l}{\textit{Architecture Design}} \\ 
    \quad DTR-B~\cite{park2023denoising} & 8.82 \\
    \rowcolor{gray!25} \quad\textbf{Switch-DiT-B} & \textbf{8.53} \\
    \arrayrulecolor{black}\bottomrule
    \end{tabular}
    }
    \vspace{-2mm}
    \caption{{\textbf{Comparative results on FFHQ.}}}
    \label{tab:ffhq}
\end{wraptable}


\begin{figure*}[t]
    \centering
    \includegraphics[width=0.32\textwidth]{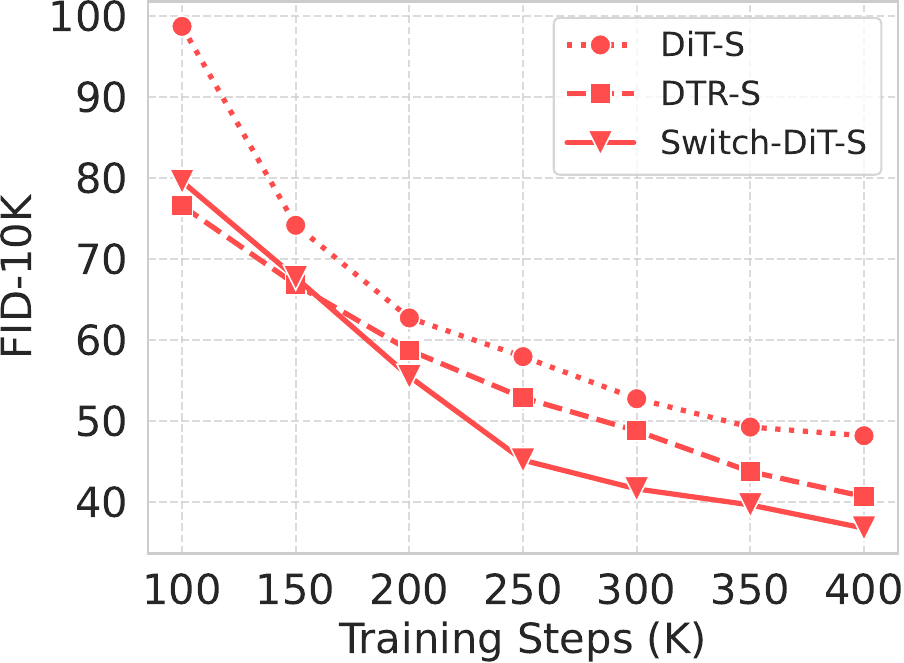}
    \includegraphics[width=0.32\textwidth]{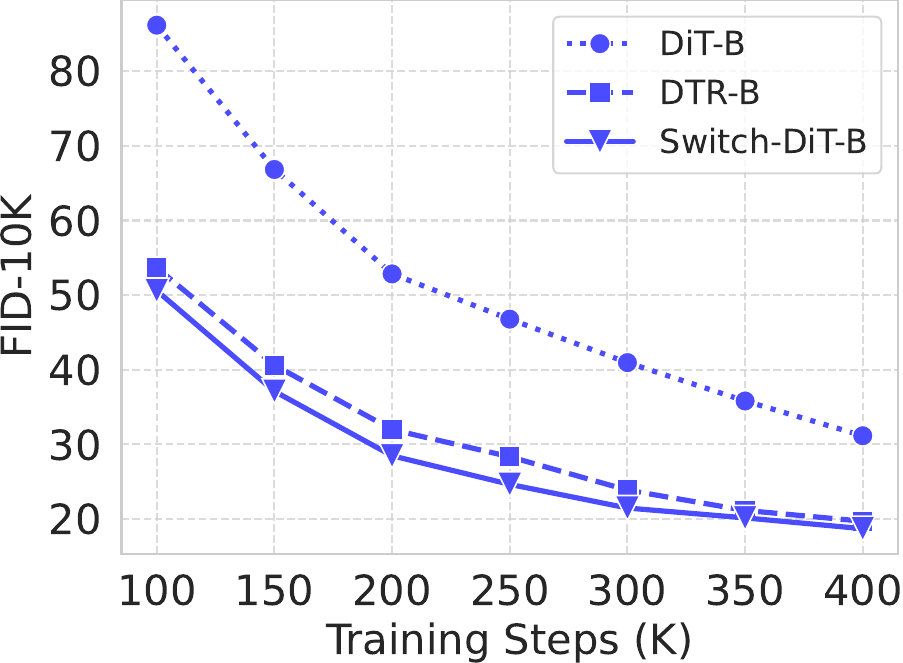}
    \includegraphics[width=0.32\textwidth]{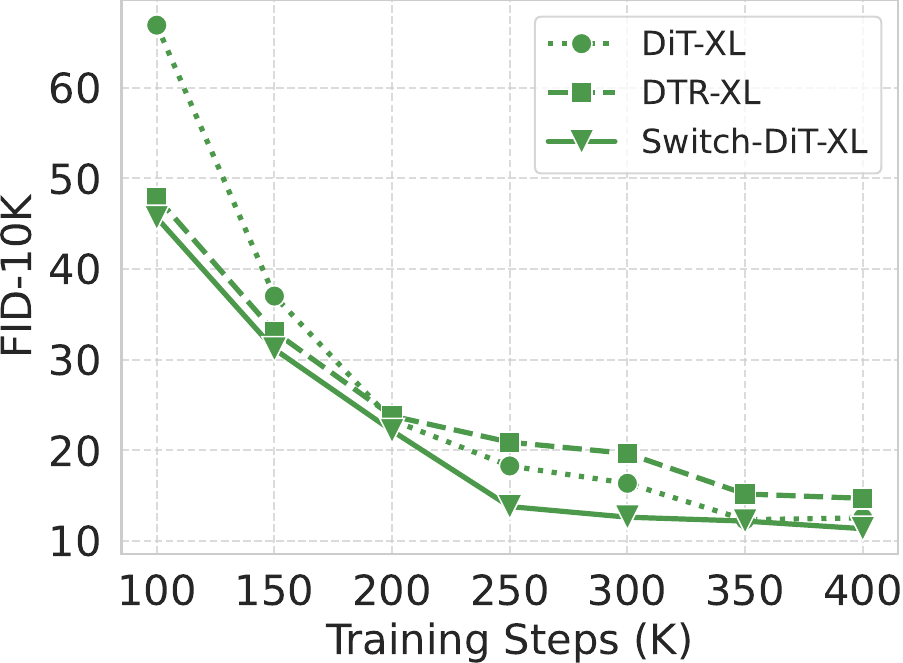}

    \caption{\textbf{Convergence comparison on ImageNet.} Switch-DiT achieves the fastest convergence rates of diffusion training across different model sizes (S, B and XL).}

    \label{fig:training_speed}
\end{figure*}

We also present comparative results with previous diffusion training methods~\cite{choi2022perception, Hang_2023_ICCV, go2024addressing} and a diffusion model architecture~\cite{park2023denoising} based on multi-task learning (MTL). Table~\ref{tab:ffhq} demonstrates the quantitative results that Switch-DiT achieves the most performance improvement compared to other MTL-based methods. In this case, we observe that explicitly isolating model parameters is more effective for handling conflicting tasks compared to relying on channel masking or optimization strategies while sharing model parameters across all timesteps.

\subsubsection{Convergence rate.} To verify the efficiency in diffusion training, we compare the convergence rate for DiT, DTR and Switch-DiT on the ImageNet dataset. The results are shown in \figref{fig:training_speed}, and we confirm that Switch-DiT consistently achieves the fastest convergence rate across all model sizes. 
Interestingly, DTR-S demonstrates superior performance during the initial stages of training, and DTR-XL exhibits faster convergence compared to DiT-XL in the early training phases. However, as the training progresses, we observe that their performance declines relative to Switch-DiT-S and DiT-XL, respectively. This indicates that channel masking effectively reduces negative transfer between conflicting tasks during the early stages of learning, while the performance degradation resulting from the loss of semantic information becomes more significant as training progresses. In contrast, our Switch-DiT represents the same inter-task relationships as DTR while establishing them through parameter isolation. This approach effectively handles conflicting tasks without losing semantic information, leading to synergizing the training of denoising tasks and consistent performance gains.

\subsection{Analysis}
\label{subsec:analysis}

\begin{figure*}[t]
    \centering
    \includegraphics[width=\linewidth]{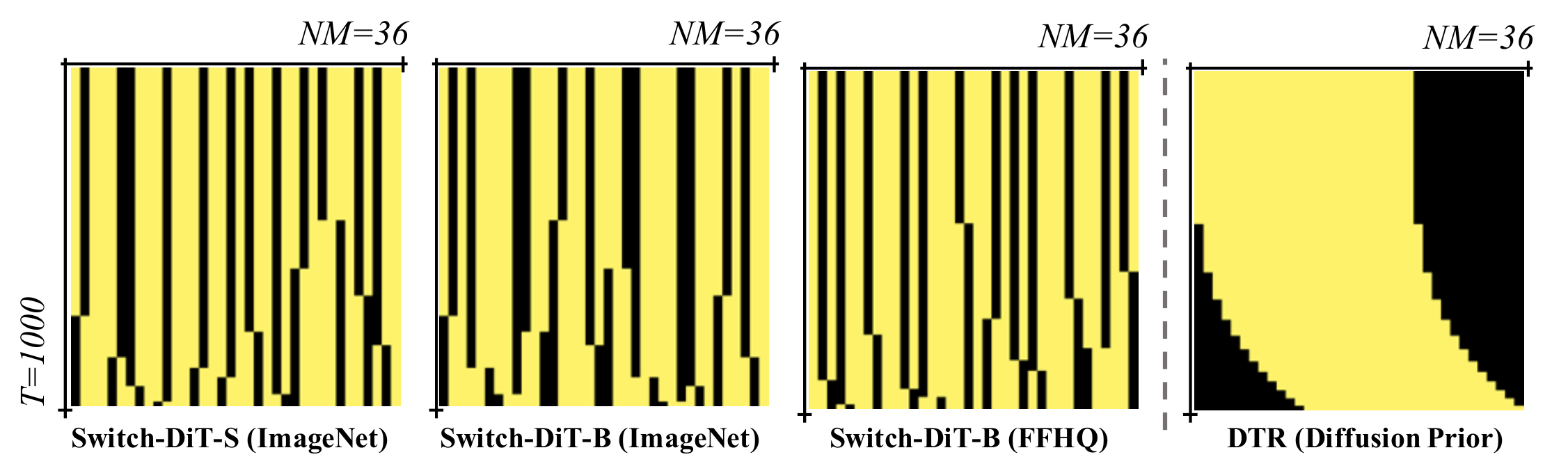}
    
    \caption{\textbf{Illustrations of stacked $\vw^{gate}_{t}$ and $\vw^{prior}_{t}$.} Even with the same diffusion prior configuration (right) as $N=12$, $M=3$, and $k=2$, we observe that the entire data pathways across all timesteps vary depending on the model size and dataset (left).}
    \label{fig:gating_variety}
\end{figure*}

\subsubsection{Gating variety.}

To demonstrate that our Switch-DiT effectively designs distinct denoising paths within the model, we visualize stacked $\vw^{gate}_{t}$ and $\vw^{prior}_{t}$ across various settings with identical configurations, as shown in \figref{fig:gating_variety}. The use of the same configuration implies employing the same gated output integration and inter-task relationships. Remarkably, we observe that denoising paths are varied across different model sizes (\ie, channel dimensions of input tokens and attention heads) on the ImageNet dataset. Furthermore, within the same model architecture, apparent discrepancies emerge between class-conditional and unconditional image generation tasks. This observation suggests that the diffusion model constructs tailored denoising paths based on different model sizes and datasets, which reflects the training signal derived from noise prediction loss.

\begin{wraptable}[4]{r}{0.33\textwidth}
    \setlength\tabcolsep{3pt}
    \centering
    \vspace{-8mm}
    \resizebox{0.33\textwidth}{!}{%
    \begin{tabular}{lc}
    \toprule
    Model & FID$\downarrow$ \\
    \toprule
    Random Allocation & 7.63 \\
    \rowcolor{gray!25}Bipartite Matching + $L_{dp}$ & \textbf{7.12} \\
    \arrayrulecolor{black}\toprule   
    \end{tabular}
    }
    \vspace{-4mm}
    \caption{\textbf{$L_{dp}$ effects.}}
    \label{tab:effect_loss}
\end{wraptable}


To further validate this characteristic, we conduct an experiment on the FFHQ dataset for \textit{Random Allocation} -- For each block, one expert is shared across all timesteps, while another expert is randomly assigned to the first $N$ columns of the activation map obtained from Eq.~\ref{eq:w_prior}, and the last one is assigned to the complementary set of the second one. As shown in Table~\ref{tab:effect_loss}, the results demonstrate that leveraging both bipartite matching and diffusion prior loss proves more effective than random allocation.

Please refer to the appendix for additional experiments and analyses on ablation studies concerning loss weight and qualitative comparisons.

\section{Conclusion}

In this work, we have presented Switch-DiT as an effective approach for leveraging inter-task relationships between conflicted denoising tasks without sacrificing semantic information. Switch-DiT enables effective parameter isolation by incorporating SMoE layers within each transformer block. The diffusion prior loss further enhances its ability to exploit detailed inter-task relationships and facilitate the rapid convergence of the EMA model.
These ensure the construction of common and task-specific denoising paths, allowing the diffusion model to construct its beneficial way of synergizing denoising tasks.
Extensive experiments have demonstrated the effectiveness of Switch-DiT in constructing tailored denoising paths across various generation scenarios. The significant improvements in image quality and convergence rate validate the efficacy of our approach.

\section*{Acknowledgements} This research was supported by Field-oriented Technology Development Project for Customs Administration through National Research Foundation of Korea (NRF) funded by the Ministry of Science \& ICT and Korea Customs Service (NRF-2021M3I1A1097906).

\bibliographystyle{splncs04}
\bibliography{egbib}

\newpage
\appendix

\section*{\centering \large Appendix}

\renewcommand{\thefigure}{\Alph{figure}}
\renewcommand{\thetable}{\Alph{table}}
\setcounter{figure}{0}
\setcounter{table}{0}

\vspace{-1mm}

We provide additional experiments in Sec.~\ref{appendix:exp} and analysis in Sec.~\ref{appendix:analysis}.

\section{Additional Experiments}
\label{appendix:exp}

\subsection{Comparison to Multi-Experts}

\begin{table*}[h]
    \centering
    \setlength\tabcolsep{2pt}
    \vspace{-8mm}
    \resizebox{0.75\textwidth}{!}{%
    \begin{tabular}{lccccc}
        \toprule
        Model & Training Steps & FID$\downarrow$  & IS$\uparrow$  & Prec$\uparrow$ & Rec$\uparrow$ \\
        \arrayrulecolor{black}\toprule
        DiT-B~\cite{peebles2022scalable} & 400K & 27.96 & 64.72 & 0.57 & 0.52 \\
        \arrayrulecolor{gray}\midrule
        Multi-experts (DiT-B $\times$ 4) & 100K $\times$ 4 & 18.01 & 78.84 & 0.61 & 0.52 \\
        DTR-B~\cite{park2023denoising} & 400K & 16.58 & 87.94 & 0.66 & \textbf{0.53} \\
        \rowcolor{gray!25}\textbf{Switch-DiT-B} & 400K & \textbf{16.21} & \textbf{88.14} & \textbf{0.68} & \textbf{0.53} \\
        \arrayrulecolor{black}\bottomrule
    \end{tabular}}
    \vspace{1mm}
    \caption{
    \textbf{Comparison to the multi-experts approach on ImageNet.}
    }
    \vspace{-10mm}
    \label{tab:multi_experts}
\end{table*}

\noindent
To further validate the architectural effectiveness of Switch-DiT, we compare it with multi-experts methods~\cite{balaji2022ediffi, Go_2023_CVPR, lee2023multi, zhang2023improving}. 
In this experiment, we train DiT-B~\cite{peebles2022scalable}, DTR-B~\cite{park2023denoising} and Switch-DiT-B for 400k iterations. For the multi-experts approach, we use four experts (\ie, DiT-B) and each expert is responsible for one of four exclusive timestep intervals in $\{1, \dots, T\}$.  Moreover, we train each expert for 100k iterations, ensuring the same training size for each denoising task. The results are shown in Table~\ref{tab:multi_experts}, and we observe that Switch-DiT is superior to the multi-experts approach for all metrics.

Key contributions to these performance improvements include that Switch-DiT incorporates detailed inter-task relationships and provides common denoising paths. For example about the multi-experts approach, if we suppose denoising tasks are grouped with uniform timestep intervals, the denoising tasks for $t=250$ and $t=249$ are grouped in the same cluster, while $t=250$ and $t=251$ are in different clusters, separating their training. This exclusive clustering falls short of capturing the inter-task relationships because it separates denoising tasks even though they are only one-timestep apart. Moreover, no parameters are learned for all denoising tasks, preventing the acquisition of shared information that may complement each task cluster. In contrast, Switch-DiT incorporates both common and task-specific denoising paths within a single diffusion model, where the degree to which these paths are shared is determined by the timestep difference and the value of the timestep itself. Also, through the well-established denoising pathways, Switch-DiT is trained to synergize denoising tasks by separating the task-specific information that can cause the negative transfer between conflicted denoising tasks, leading to significant performance improvements.

\subsection{Comparison for Model Configuration \textit{Large}}

\begin{wraptable}[6]{r}{0.51\textwidth}
    \centering
    \setlength\tabcolsep{5pt}
    \vspace{-8mm}
    \resizebox{0.51\textwidth}{!}{%
    \begin{tabular}{lcccc}
        \toprule
        Model & FID$\downarrow$  & IS$\uparrow$  & Prec$\uparrow$ & Rec$\uparrow$ \\
        \arrayrulecolor{black}\toprule
        DiT-L~\cite{peebles2022scalable} & 12.59 & 134.60 & 0.73 & 0.49 \\
        DTR-L~\cite{park2023denoising} & 8.90 & 156.48 & 0.77 & \textbf{0.51} \\
        \rowcolor{gray!25}\textbf{Switch-DiT-L} & \textbf{8.78} & \textbf{162.97} & \textbf{0.78} & 0.50 \\
        \bottomrule
    \end{tabular}}
    \vspace{-3.5mm}
    \caption{
    \textbf{Comparative results of model configuration \textit{Large} on ImageNet.}
    }
    \label{tab:config_l}
\end{wraptable}

We provide results for the model configuration \textit{Large} in Table~\ref{tab:config_l}. Remarkably, Switch-DiT-L improves DiT-L and DTR-L, further validating the consistent performance improvement with respect to model sizes.

\begin{wraptable}[7]{r}{0.37\textwidth}
    \centering
    \setlength\tabcolsep{7pt}
    \vspace{-6mm}
    \resizebox{0.3\textwidth}{!}{%
    \begin{tabular}{lc}
        \toprule
        Method & FID-10K$\downarrow$ \\
        \toprule
        DiT-S & 92.07 \\
        DTR-S & 42.24 \\
        \arrayrulecolor{gray}\midrule
        Switch-DiT-S & 34.39 \\
        \arrayrulecolor{black}\bottomrule
    \end{tabular}}
    \vspace{-2mm}
    \caption{
    \textbf{Results on FFHQ.}
    }
    \label{tab:high_ffhq}
\end{wraptable}

\subsection{High-resolution.}
We further validate the effectiveness of our method on the FFHQ dataset with a higher resolution of 512 $\times$ 512. We trained DiT-S, DTR-S, and Switch-DiT-S for 100k iterations. As shown in \Cref{tab:high_ffhq}, our Switch-DiT achieves significant improvements compared to previous methods. 

\begin{figure*}[t!]
    \centering
        \begin{tabular}{@{\hspace{2pt}}c@{\hspace{2pt}}c@{\hspace{2pt}}c@{\hspace{2pt}}c@{\hspace{2pt}}c@{\hspace{2pt}}c@{\hspace{2pt}}c}
        \toprule
        \multicolumn{7}{l}{\scriptsize\textbf{Class-conditional Image Generation on ImageNet~\cite{deng2009imagenet}}} \\
        \toprule
        \raisebox{0.06\textwidth}{\rotatebox{90}{\scriptsize S}} 
        \adjincludegraphics[clip,width=0.125\textwidth,trim={0 0 0 0}]{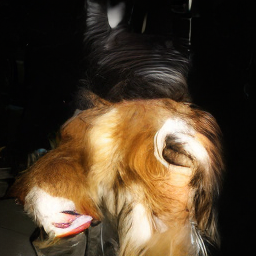} &
        \adjincludegraphics[clip,width=0.125\textwidth,trim={0 0 0 0}]{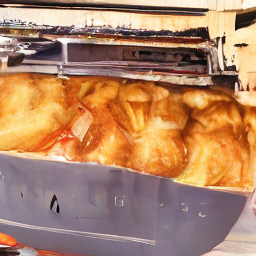} &
        \adjincludegraphics[clip,width=0.125\textwidth,trim={0 0 0 0}]{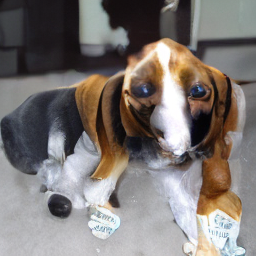} &
        \adjincludegraphics[clip,width=0.125\textwidth,trim={0 0 0 0}]{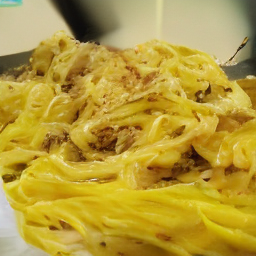} &
        \adjincludegraphics[clip,width=0.125\textwidth,trim={0 0 0 0}]{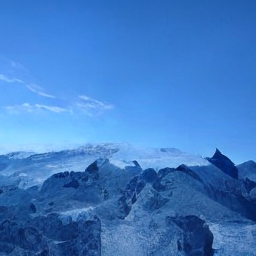} &
        \adjincludegraphics[clip,width=0.125\textwidth,trim={0 0 0 0}]{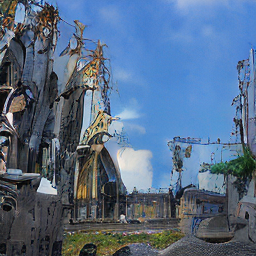} &
        \adjincludegraphics[clip,width=0.125\textwidth,trim={0 0 0 0}]{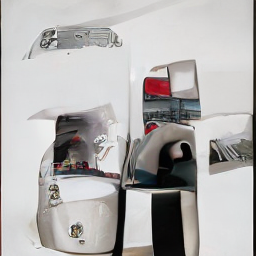} \\
        
        \raisebox{0.06\textwidth}{\rotatebox{90}{\scriptsize B}} 
        \adjincludegraphics[clip,width=0.125\textwidth,trim={0 0 0 0}]{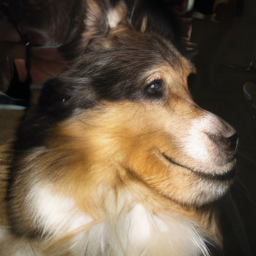} &
        \adjincludegraphics[clip,width=0.125\textwidth,trim={0 0 0 0}]{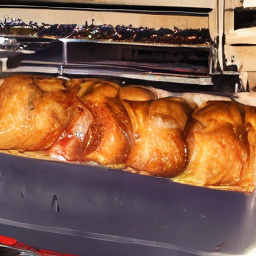} &
        \adjincludegraphics[clip,width=0.125\textwidth,trim={0 0 0 0}]{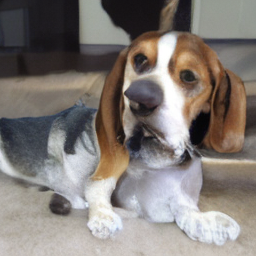} &
        \adjincludegraphics[clip,width=0.125\textwidth,trim={0 0 0 0}]{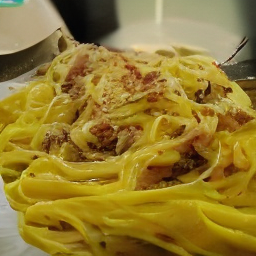} &
        \adjincludegraphics[clip,width=0.125\textwidth,trim={0 0 0 0}]{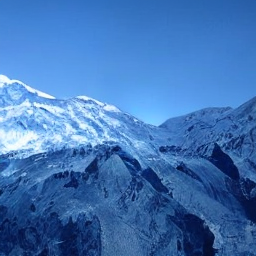} &
        \adjincludegraphics[clip,width=0.125\textwidth,trim={0 0 0 0}]{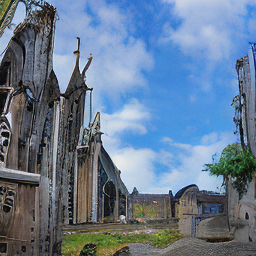} &
        \adjincludegraphics[clip,width=0.125\textwidth,trim={0 0 0 0}]{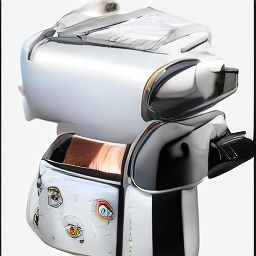} \\
        
        \raisebox{0.05\textwidth}{\rotatebox{90}{\scriptsize XL}} 
        \adjincludegraphics[clip,width=0.125\textwidth,trim={0 0 0 0}]{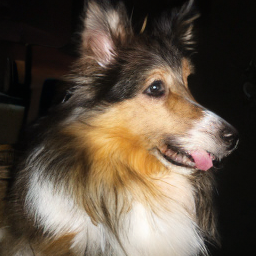} &
        \adjincludegraphics[clip,width=0.125\textwidth,trim={0 0 0 0}]{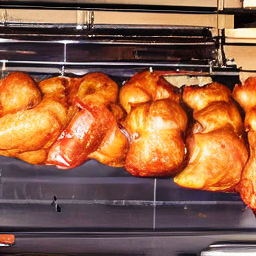} &
        \adjincludegraphics[clip,width=0.125\textwidth,trim={0 0 0 0}]{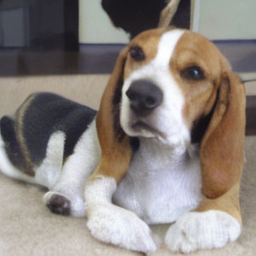} &
        \adjincludegraphics[clip,width=0.125\textwidth,trim={0 0 0 0}]{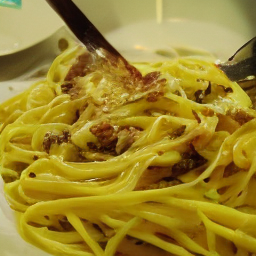} &
        \adjincludegraphics[clip,width=0.125\textwidth,trim={0 0 0 0}]{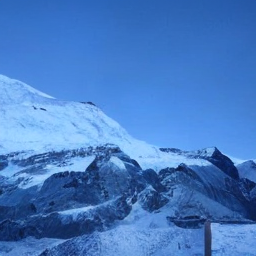} &
        \adjincludegraphics[clip,width=0.125\textwidth,trim={0 0 0 0}]{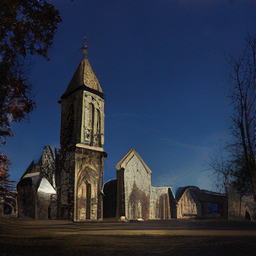} &
        \adjincludegraphics[clip,width=0.125\textwidth,trim={0 0 0 0}]{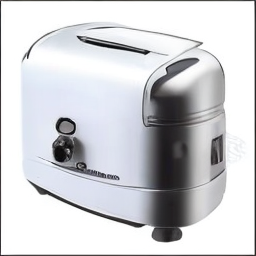} \\
        \quad{\scriptsize Shetland} & {\scriptsize Rotisserie} & {\scriptsize Beagle} & {\scriptsize Carbonara} & {\scriptsize Alp} & {\scriptsize Church} & {\scriptsize Toaster} \\
        \bottomrule
    \end{tabular}
    \caption{
        \textbf{Qualitative results of Switch-DiT on ImageNet.}
    }
    \label{fig:qualitative_scalability}
\end{figure*}

\section{Additional Ananysis}
\label{appendix:analysis}

\subsection{Scalability}

\begin{wrapfigure}[10]{r}{0.35\textwidth}
    \centering
    \vspace{-6mm}
    \includegraphics[width=0.95\linewidth]{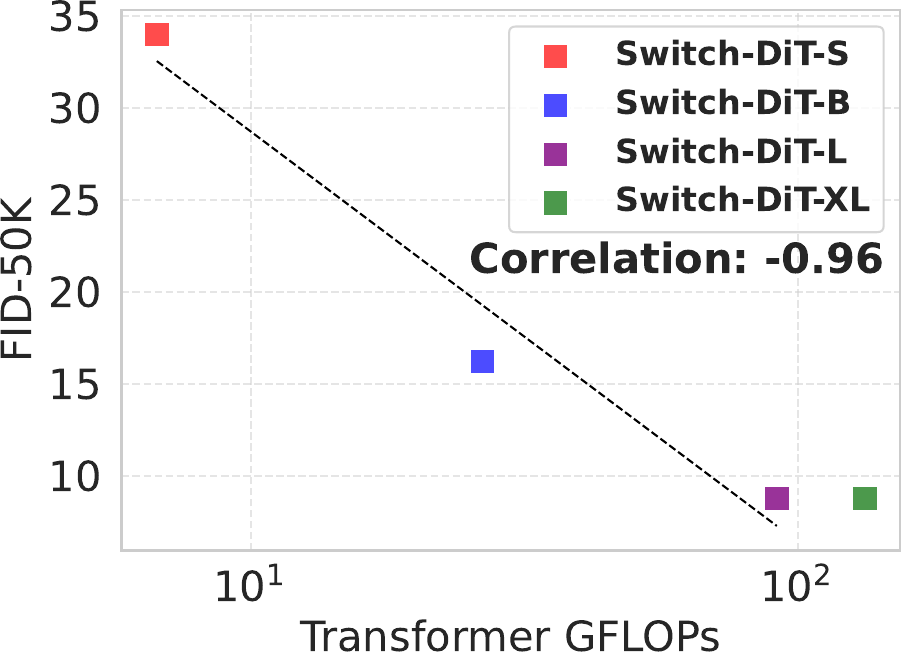}
    \vspace{-3mm}
    \caption{\textbf{Switch-DiT scaling behavior on ImageNet.}}
    \label{fig:correlation_switch}
\end{wrapfigure}

Figure~\ref{fig:correlation_switch} shows the correlation between GFLOPs and the FID-50K score for the Switch-DiT. Similar to DiT~\cite{peebles2022scalable}, we also observe a strong correlation from Switch-DiT-S to Switch-DiT-L, confirming its scaling behavior. Notably, in terms of the FID score, Switch-DiT-L outperforms DiT-XL even in smaller model configurations (9.40 vs. 8.78), further demonstrating its scalability with the lower FID score of 8.76 achieved by Switch-DiT-XL. This implies that our SMoE-based architectural design is more effective for scaling diffusion models than simply increasing the hidden size and the number of transformer blocks.

Also, Fig.~\ref{fig:qualitative_scalability} illustrates the impacts of scaling model size on sample quality. The results demonstrate that larger model sizes for Switch-DiT consistently produce better images across multiple classes.

\begin{figure*}[t!]
    \centering
    \includegraphics[width=0.95\linewidth]{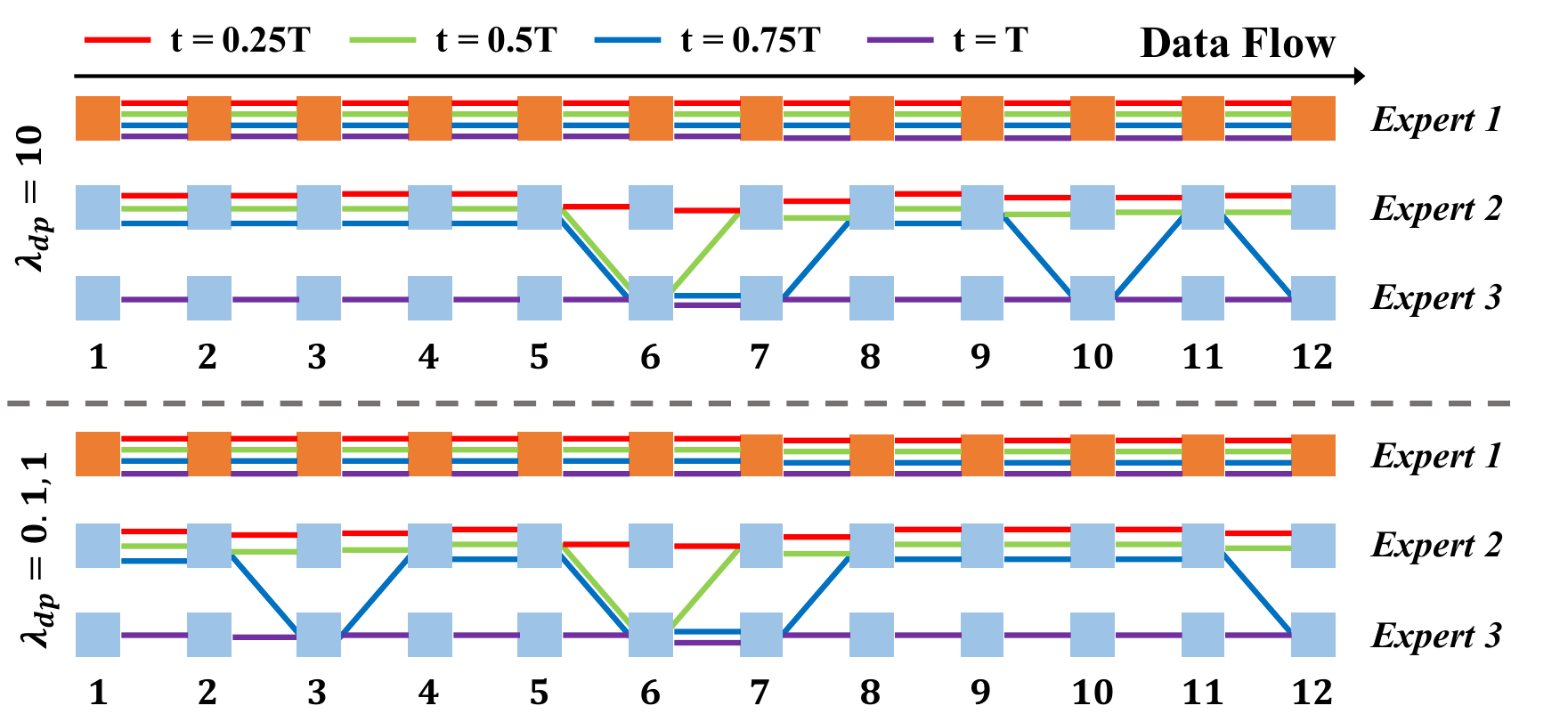}
    \caption{\textbf{Denoising path across $\lambda_{dp}$ on FFHQ.} We show the denoising paths of Switch-DiT-B for four timesteps ($0.25T, 0.5T, 0.75T, T$). We sort the expert index for simplicity, and experts in orange are common experts shared across all timesteps.}
    \vspace{-3mm}
    \label{fig:denoising_path}
\end{figure*}

\begin{figure*}[t!]
    \centering
        \begin{tabular}{@{\hspace{2pt}}c@{\hspace{2pt}}c@{\hspace{2pt}}c@{\hspace{2pt}}c@{\hspace{2pt}}c@{\hspace{2pt}}c@{\hspace{2pt}}c}
        \toprule
        \multicolumn{7}{l}{\scriptsize\textbf{Unconditional Image Generation on FFHQ~\cite{karras2019style} (Model Configuration: B)}} \\
        \toprule
        
        \raisebox{0.04\textwidth}{\rotatebox{90}{\scriptsize DiT}} 
        \adjincludegraphics[clip,width=0.125\textwidth,trim={0 0 0 0}]{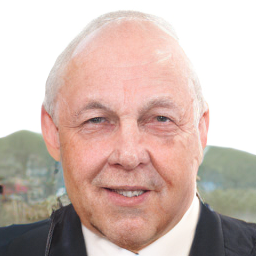} &
        \adjincludegraphics[clip,width=0.125\textwidth,trim={0 0 0 0}]{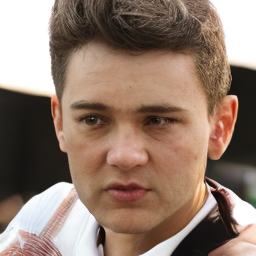} &
        \adjincludegraphics[clip,width=0.125\textwidth,trim={0 0 0 0}]{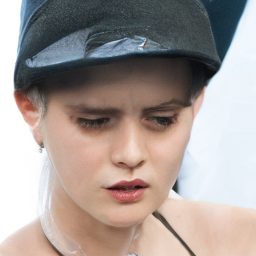} &
        \adjincludegraphics[clip,width=0.125\textwidth,trim={0 0 0 0}]{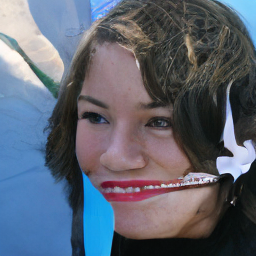} &
        \adjincludegraphics[clip,width=0.125\textwidth,trim={0 0 0 0}]{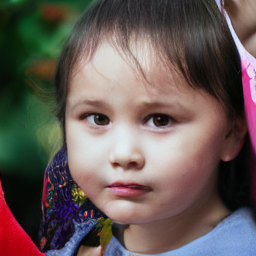} &
        \adjincludegraphics[clip,width=0.125\textwidth,trim={0 0 0 0}]{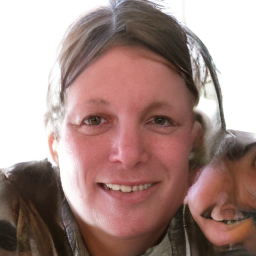} &
        \adjincludegraphics[clip,width=0.125\textwidth,trim={0 0 0 0}]{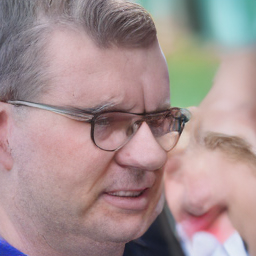} \\
        
        \raisebox{0.04\textwidth}{\rotatebox{90}{\scriptsize DTR}} 
        \adjincludegraphics[clip,width=0.125\textwidth,trim={0 0 0 0}]{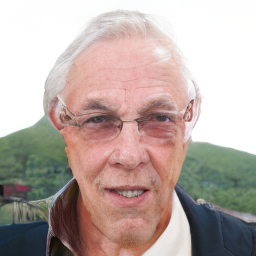} &
        \adjincludegraphics[clip,width=0.125\textwidth,trim={0 0 0 0}]{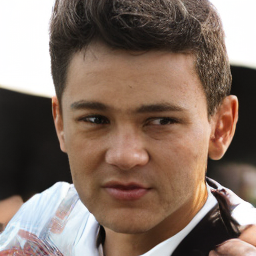} &
        \adjincludegraphics[clip,width=0.125\textwidth,trim={0 0 0 0}]{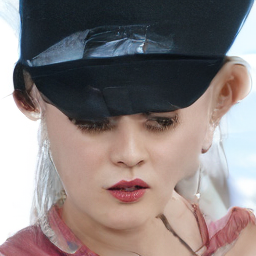} &
        \adjincludegraphics[clip,width=0.125\textwidth,trim={0 0 0 0}]{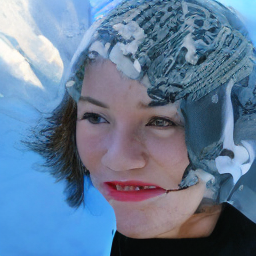} &
        \adjincludegraphics[clip,width=0.125\textwidth,trim={0 0 0 0}]{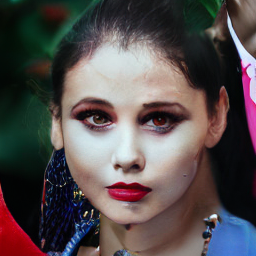} &
        \adjincludegraphics[clip,width=0.125\textwidth,trim={0 0 0 0}]{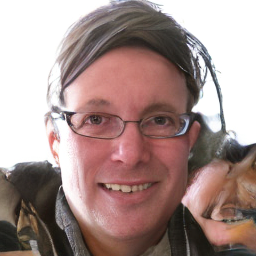} &
        \adjincludegraphics[clip,width=0.125\textwidth,trim={0 0 0 0}]{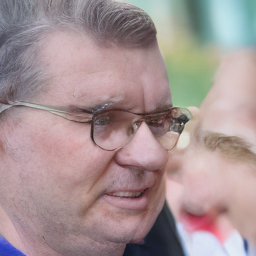} \\
        
        \raisebox{0.07\height}{\rotatebox{90}{\scriptsize Switch-DiT}} 
        \adjincludegraphics[clip,width=0.125\textwidth,trim={0 0 0 0}]{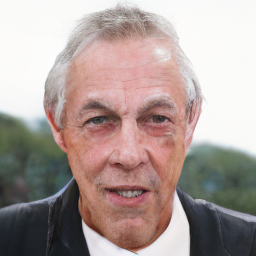} &
        \adjincludegraphics[clip,width=0.125\textwidth,trim={0 0 0 0}]{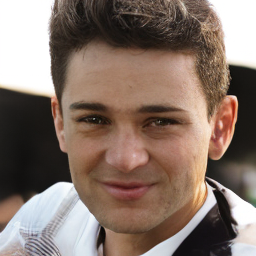} &
        \adjincludegraphics[clip,width=0.125\textwidth,trim={0 0 0 0}]{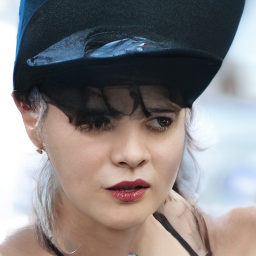} &
        \adjincludegraphics[clip,width=0.125\textwidth,trim={0 0 0 0}]{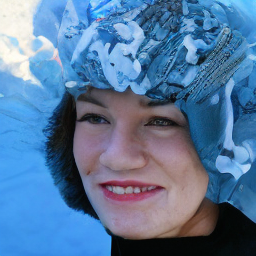} &
        \adjincludegraphics[clip,width=0.125\textwidth,trim={0 0 0 0}]{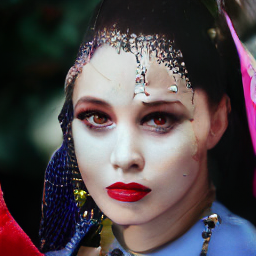} &
        \adjincludegraphics[clip,width=0.125\textwidth,trim={0 0 0 0}]{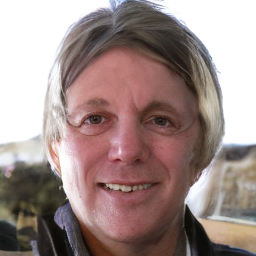} &
        \adjincludegraphics[clip,width=0.125\textwidth,trim={0 0 0 0}]{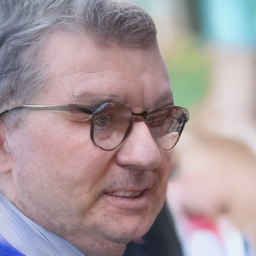} \\
        \bottomrule
    \end{tabular}
    \begin{tabular}{@{\hspace{2pt}}c@{\hspace{2pt}}c@{\hspace{2pt}}c@{\hspace{2pt}}c@{\hspace{2pt}}c@{\hspace{2pt}}c@{\hspace{2pt}}c}
        \multicolumn{7}{l}{\scriptsize\textbf{Class-conditional Image Generation on ImageNet~\cite{deng2009imagenet} (Model Configuration: XL)}} \\
        \arrayrulecolor{gray}\midrule
        
        \raisebox{0.04\textwidth}{\rotatebox{90}{\scriptsize DiT}} 
        \adjincludegraphics[clip,width=0.125\textwidth,trim={0 0 0 0}]{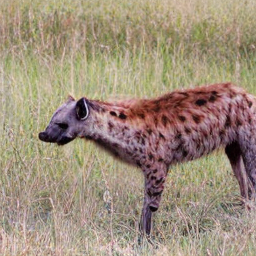} &
        \adjincludegraphics[clip,width=0.125\textwidth,trim={0 0 0 0}]{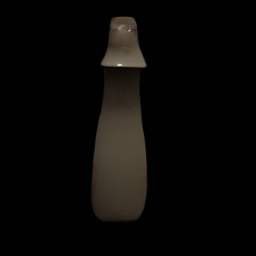} &
        \adjincludegraphics[clip,width=0.125\textwidth,trim={0 0 0 0}]{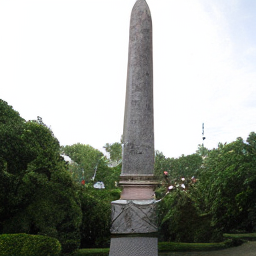} &
        \adjincludegraphics[clip,width=0.125\textwidth,trim={0 0 0 0}]{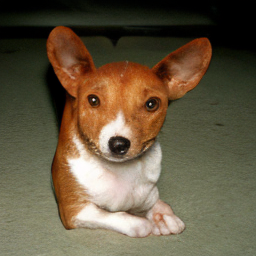} &
        \adjincludegraphics[clip,width=0.125\textwidth,trim={0 0 0 0}]{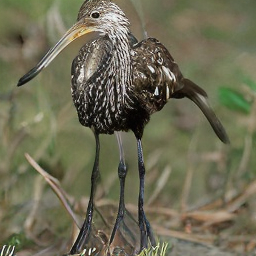} &
        \adjincludegraphics[clip,width=0.125\textwidth,trim={0 0 0 0}]{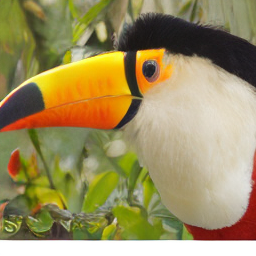} &
        \adjincludegraphics[clip,width=0.125\textwidth,trim={0 0 0 0}]{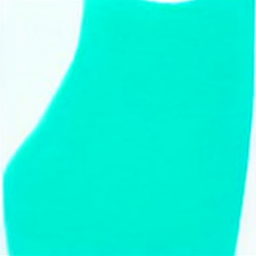} \\
        
        \raisebox{0.04\textwidth}{\rotatebox{90}{\scriptsize DTR}} 
        \adjincludegraphics[clip,width=0.125\textwidth,trim={0 0 0 0}]{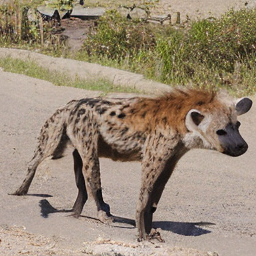} &
        \adjincludegraphics[clip,width=0.125\textwidth,trim={0 0 0 0}]{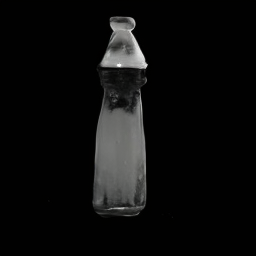} &
        \adjincludegraphics[clip,width=0.125\textwidth,trim={0 0 0 0}]{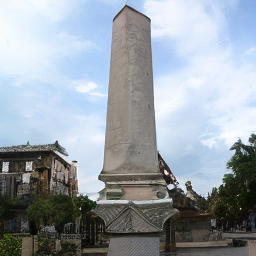} &
        \adjincludegraphics[clip,width=0.125\textwidth,trim={0 0 0 0}]{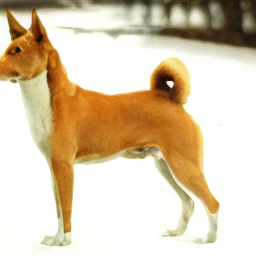} &
        \adjincludegraphics[clip,width=0.125\textwidth,trim={0 0 0 0}]{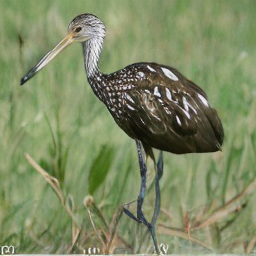} &
        \adjincludegraphics[clip,width=0.125\textwidth,trim={0 0 0 0}]{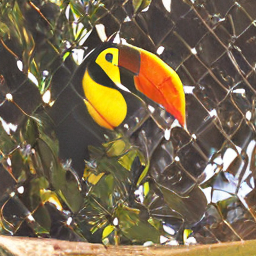} &
        \adjincludegraphics[clip,width=0.125\textwidth,trim={0 0 0 0}]{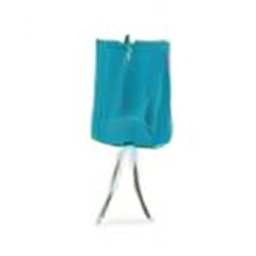} \\
        
        \raisebox{0.07\height}{\rotatebox{90}{\scriptsize Switch-DiT}} 
        \adjincludegraphics[clip,width=0.125\textwidth,trim={0 0 0 0}]{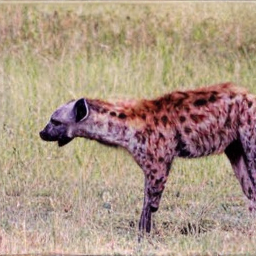} &
        \adjincludegraphics[clip,width=0.125\textwidth,trim={0 0 0 0}]{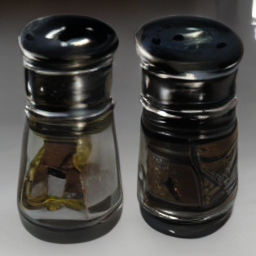} &
        \adjincludegraphics[clip,width=0.125\textwidth,trim={0 0 0 0}]{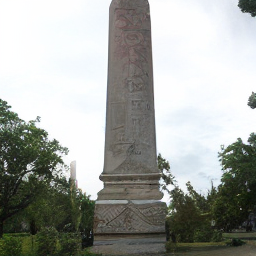} &
        \adjincludegraphics[clip,width=0.125\textwidth,trim={0 0 0 0}]{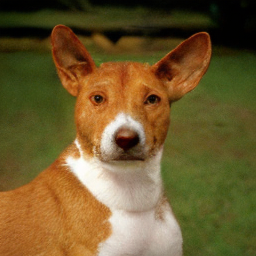} &
        \adjincludegraphics[clip,width=0.125\textwidth,trim={0 0 0 0}]{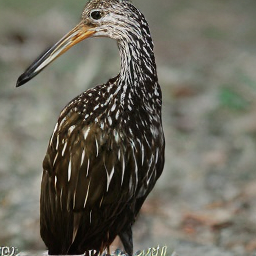} &
        \adjincludegraphics[clip,width=0.125\textwidth,trim={0 0 0 0}]{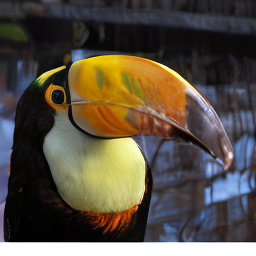} &
        \adjincludegraphics[clip,width=0.125\textwidth,trim={0 0 0 0}]{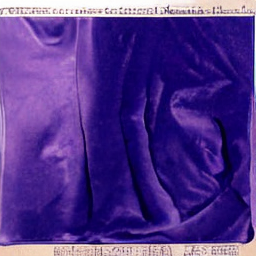} \\
        \quad{\scriptsize Hyena} & {\scriptsize Saltshaker} & {\scriptsize Obelisk} & {\scriptsize Basenji} & {\scriptsize Limpkin} & {\scriptsize Toucan} & {\scriptsize Velvet} \\
        \arrayrulecolor{black}\bottomrule
    \end{tabular}
    
    \vspace{\abovetabcapmargin}
    \caption{
        \textbf{Qualitative Comparison on FFHQ and ImageNet datasets.}
    }
    \vspace{-2.5mm}
    \label{fig:qualitative_comparison}
\end{figure*}

\subsection{Impacts on Loss Weight}

\begin{wraptable}[6]{r}{0.21\textwidth}
    \setlength\tabcolsep{12pt}
    \centering
    \vspace{-8mm}
    \resizebox{0.21\textwidth}{!}{%
    \begin{tabular}{lc}
    \toprule
    $\lambda_{dp}$ & FID$\downarrow$ \\
    \toprule
    0.1 & 7.42 \\
    \rowcolor{gray!25} \textbf{1.0} & \textbf{7.12} \\
    10.0 & 7.58 \\
    \arrayrulecolor{black}\bottomrule
    \end{tabular}
    }
    \vspace{-3.2mm}
    \caption{{\textbf{$\lambda_{dp}$ effects on FFHQ.}}}
    \label{tab:loss_weight}
\end{wraptable}

We investigate the impact of varying $\lambda_{dp}$.
Table~\ref{tab:loss_weight} presents the ablative results on the FFHQ dataset, and we observe that the performance peaks at $\lambda_{dp}=1$.
Also, the performance differences are not significantly influenced by changes in $\lambda_{dp}$, suggesting that the diffusion prior loss provides robust supervision on the gating output.
However, we observe that the construction of the denoising path within the diffusion model changes as our gating networks are implicitly trained via the noise prediction loss $\gL_{noise}$, and directly supervised by the diffusion prior loss $\gL_{dp}$.

Figure~\ref{fig:denoising_path} shows the denoising paths of Switch-DiT-B on the FFHQ dataset. Interestingly, denoising paths for $\lambda_{dp} = 0.1$ and $\lambda_{dp} = 1$ are the same, while those for $\lambda_{dp} = 10$ exhibits different paths. This discrepancy arises because the gating network does not fully reflect the training signal from $\gL_{noise}$ when $\lambda_{dp}=10$. Instead, the determination of denoising paths is primarily influenced by the inherent randomness in the bipartite matching process. Additionally, we observe that the denoising paths converge within 500 iterations for $\lambda_{dp} = 1$ and $\lambda_{dp} = 10$, whereas it takes thousands of iterations for $\lambda_{dp} = 0.1$. 
The slow convergence is attributed to the failure to promptly reflect inter-task relationships, thus leading to ineffective management of negative transfers between conflicting tasks during the early stages of learning. In contrast, when $\lambda_{dp} = 1$, gating networks properly reflect training signals from noise prediction loss and inter-task relationships.

\subsection{Qualitative comparisons.}

Figure~\ref{fig:qualitative_comparison} shows the qualitative comparison of DiT~\cite{peebles2022scalable}, DTR~\cite{park2023denoising} and Switch-DiT. The results demonstrate that Switch-DiT generates more realistic images, further verifying the effectiveness of our proposed method.

\section{Discussion}

\subsubsection{Ethics Statement}
Our approach is one of the generative models, and it may carry significant societal implications, particularly in applications like deep fakes and addressing biased data.

\vspace{-2mm}
\subsubsection{Limiatations and Future Works}

In this work, we have employed the most parameter-efficient SMoE configuration, to effectively synergize denoising tasks within the diffusion model. While we have adopted the routing policy from DTR, which uses the same policy across different model sizes and datasets, this fixed routing policy may not fully capture the nuanced inter-task relationships. Therefore, there is potential for enhancing our approach by tailoring the inter-task relationships to specific generation scenarios. Additionally, further exploration of configurations such as the number of experts in each transformer block and the TopK value can also facilitate the scalability of transformer-based diffusion models, such as the evolution of large language models with SMoE layers.

\end{document}